\documentclass[sigconf,natbib=true]{acmart}

\AtBeginDocument{%
  \providecommand\BibTeX{{%
    \normalfont B\kern-0.5em{\scshape i\kern-0.25em b}\kern-0.8em\TeX}}}

\setcopyright{acmlicensed}
\copyrightyear{2018}
\acmYear{2018}
\acmDOI{XXXXXXX.XXXXXXX}

\acmConference[Conference acronym 'XX]{Make sure to enter the correct
  conference title from your rights confirmation emai}{June 03--05,
  2018}{Woodstock, NY}
\acmISBN{978-1-4503-XXXX-X/18/06}
\usepackage{amsfonts} 
\usepackage{todonotes}
\usepackage{xspace}
\usepackage{multirow}
\newcommand{\ch}{\mathbf{h}\xspace}
\newcommand{\cm}{\mathbf{m}\xspace}

\newcommand{\x}{\mathbf{x}\xspace}

\newcommand{\cH}{\mathbf{H}\xspace}

\newcommand{\cmsg}{\mathbf{msg}\xspace}

\newcommand{\CG}{\mathcal{G}\xspace}
\newcommand{\cw}{\mathbf{w}\xspace}

\newcommand{\CN}{\mathcal{N}\xspace}
\newcommand{\CV}{\mathcal{V}\xspace}
\newcommand{\CI}{\mathcal{I}\xspace}

\newcommand{\CE}{\mathcal{E}\xspace}
\newcommand{\CX}{\mathcal{X}\xspace}

\newcommand{\CU}{\mathcal{U}\xspace}

 \newtheorem{defn}{\textbf{Definition}}
  \newtheorem{prob}{\textbf{Problem}}

\def\tgnn{\textsc{Tgnn}\xspace}
\def\gm{\textsc{GraphMixer}\xspace}

\def\gnn{\textsc{GNN}\xspace}
\def\tgat{\textsc{TGAT}\xspace}
\def\gcn{\textsc{GCN}\xspace}
\def\tgn{\textsc{TGN}\xspace}
\def\dygformer{\textsc{Dygformer}\xspace}

\def\fgat{\textsc{F-GAT}\xspace}
\def\mtgn{\textsc{Mintt}\xspace}

\def\ngcf{\textsc{NGCF}\xspace}
\def\lightgcn{\textsc{LightGCN}\xspace}

\newcommand{\namemodelnew}{\textsc{Mintt}\xspace}
\usepackage{paralist}
\usepackage{subcaption}
\begin{document}

\title{A Transfer Framework for Enhancing Temporal Graph Learning in Data-Scarce Settings}

\author{Sidharth Agarwal}
\authornote{Both authors contributed equally to this research.}
\affiliation{%
  \institution{Indian Institute of Technology, Delhi}
  \city{New Delhi}
  \country{India}
}
\email{sidharthagarwal2001@gmail.com}

\author{Tanishq Dubey}
\authornotemark[1]
\affiliation{%
  \institution{Indian Institute of Technology, Delhi}
  \city{New Delhi}
  \country{India}
}
\email{tanishqd1420@gmail.com}

\author{Shubham Gupta}
\affiliation{%
  \institution{Indian Institute of Technology, Delhi}
  \city{New Delhi}
  \country{India}
}
\email{shubham.gupta@cse.iitd.ac.in}

\author{Srikanta Bedathur}
\affiliation{%
  \institution{Indian Institute of Technology, Delhi}
 \city{New Delhi}
  \country{India}
}
\email{srikanta@cse.iitd.ac.in}
\begin{abstract}

Dynamic interactions between entities are prevalent in domains like social platforms, financial systems, healthcare, and e-commerce. These interactions can be effectively represented as time-evolving graphs, where predicting future connections is a key task in applications such as recommendation systems. Temporal Graph Neural Networks (TGNNs) have achieved strong results for such predictive tasks but typically require extensive training data, which is often limited in real-world scenarios. One approach to mitigating data scarcity is leveraging pre-trained models from related datasets. However, direct knowledge transfer between TGNNs is challenging due to their reliance on node-specific memory structures, making them inherently difficult to adapt across datasets.

To address this, we introduce a novel transfer approach that disentangles node representations from their associated features through a structured bipartite encoding mechanism. This decoupling enables more effective transfer of memory components and other learned inductive patterns from one dataset to another. Empirical evaluations on real-world benchmarks demonstrate that our method significantly enhances TGNN performance in low-data regimes, outperforming non-transfer baselines by up to 56\% and surpassing existing transfer strategies by 36\%

\end{abstract}




\maketitle

\section{Introduction and Related Work}
\begin{figure}[t]
    \centering
    \includegraphics[scale=0.4]{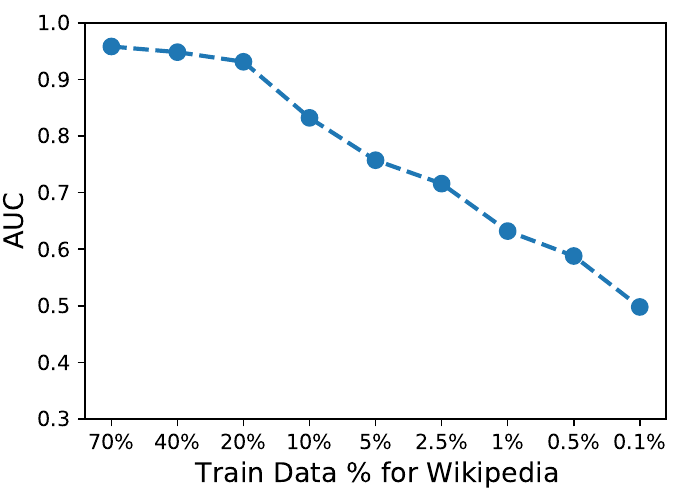}
    \caption{Performance degradation of \tgnn on future link prediction task as training data is progressively reduced on \textsc{wiki-edit}.}
\vspace{-0.2in}
    \label{fig:data_scarcity_wiki_edit}
\end{figure}

 Graph Neural Networks (\gnn) are widely used in applications like node classification \cite{graphsage,gat,gcn,gin}, link prediction \cite{rgcn,chamberlain2023graph}, traffic prediction \cite{10.5555/3304222.3304273},  combinatorial optimization \cite{Schuetz_2022}, and drug discovery \cite{graphrnn,netwalk,graphvae}. However, they assume the underlying relationships between nodes to be static, while real-world systems, e.g., interaction networks, are dynamic. Thus, it becomes imperative to learn dynamic inductive patterns. Recently, temporal graph neural networks (\tgnn) \cite{jodie,tgat,tgn,pint,caw,HTNE} have proved very effective in downstream tasks like future link prediction and dynamic node classification by capturing structural and temporal characteristics. The superior performance of \tgnn is conditioned on the availability of large-scale  {timestamped interactions, which are unavailable when a dynamic system is at an early phase in many real-world settings.  {This is true in general for all neural models \cite{NIPS2017_3f5ee243,329294}}. We demonstrate this effect on \tgnn by simulating data scarcity, reducing the training dataset of \textsc{wiki-edit}\cite{jodie} by selecting only the earliest interactions sorted by the timestamp}, as shown in Figure \ref{fig:data_scarcity_wiki_edit}. When trained with 5--10\% of  {initial} data, the performance of the model declines by more than 30\%.

In real-world applications, an online delivery enterprise entering a new city might face insufficient data to train a \tgnn-based restaurant recommendation system. Transfer learning \cite{10.5555/2969033.2969197} addresses this by leveraging historical data from existing cities. The model is initially trained on the source city's data and then fine-tuned on limited data from the target city before deployment. Early \tgnn like \tgat \cite{tgat}, mainly consisted of the temporal attention layer  {which applies local aggregator operator on \textit{past temporal neighbors} processing their hidden representation, attributes, and time-stamp information to compute node embedding at any given time $t$}. Model transfers thus involved reusing the aggregators' weights and fine-tuning the target dataset.

Recent advancements in  \tgnn, such as state-of-art methods \cite{tgn, pint} have introduced a novel memory module that compresses a node's historical interactions into a memory vector using \textit{recurrent neural network}.  This memory is passed to the local aggregator module in temporal convolution to compute dynamic node embedding. The memory vector of a node is updated via a {GRU} cell \cite{gru} during an interaction event and stores the long-term dependencies, significantly enhancing the capabilities and observed performance of the temporal graph attention networks.




Transferring these recent \tgnn involves migrating the weights of local aggregator module, {GRU} module, and current memory vectors to fully leverage the source graph's patterns. Since the memory is node-specific, i.e., transductive, a direct model transfer is not feasible. Rather, the ideal solution requires mapping nodes in the target graph to the nodes in the source graph and utilizing the use of the corresponding memory during transfer. 

\subsection{Existing works and their limitations}
\label{subsection:existing_work}
 {Currently, there doesn't exist any transfer framework for temporal graph networks (\tgnn). We are the first to motivate, define, and propose a solution for transferring memory-based \tgnn. However, there are methods for transfer learning over \textit{Graph Neural Networks} ({\gnn}) which operate on static graphs.  These works have the following limitations:}
\begin{itemize}
\item {\textbf{Limited to non-temporal graphs:}} 
Existing works \cite{gfewshot,srcreweights,yao2020learning,kooverjee2022investigating,grl} are primarily designed for static graphs. Since edges in static graphs are fixed, there is no need for supporting memory for every node, as their behavior patterns are not dynamic.
\item {\textbf{Limited to cold start recommendations in source graph:}} Existing cold-start recommendation \gnn based transfer methods \cite{cold_start_gnn,pre_train_gnn_cold_start,mamo} primarily simulates cold-start scenario during training given the data and train a robust model using meta-learning based framework. This trained model generates representations for new nodes in the same source graph, while our emphasis is on learning representations for nodes on the target graph. 


\item {\textbf{Require joint training of source and target datasets:}} Existing works based on network alignment \cite{grl,10.5555/3367243.3367352} learn a joint representation of nodes in both source \& target datasets by applying regularization to be in same representation space. This has limited applicability as it requires joint re-training for every target while we focus on the transfer of a trained \tgnn model to any target. 

By focusing on learning the representation of nodes on the target graph and avoiding joint training, our proposed model can adapt better to the specific structural \& temporal features of the target domain. Please refer to section \ref{app:related_works} for detailed literature review.

\end{itemize}
\vspace{-0.16in}
\subsection{Contribution}
This work proposes a \textbf{M}emory \textbf{In}ductive \textbf{T}ransfer framework for \textbf{T}emporal graph neural networks (\namemodelnew) from source temporal interaction dataset to target dataset in a scarcity setting, which addresses the limitations of the prior work highlighted above. Following are the novel contributions of this work:
\begin{itemize}
    \item {\textbf{Transfer Framework for \tgnn:}} As highlighted earlier, transferring \tgnn requires weight transfer along with memory transfer. Since memory is transductive, we propose to learn the universal node embedding model by transforming source and target temporal graphs into a graph containing original graph nodes and virtual feature nodes.  This transformation allows the utilize common attributes between source and target. A novel message-passing framework on this transformation enables mapping nodes in the target graph to nodes in the source graph, facilitating memory transfer. As per our knowledge, this is the first work to facilitate the transfer of temporal graph neural networks from source to scarce target temporal interaction graphs.

    \item {\textbf{Empirical Evaluation:}} We evaluate our proposed transfer method on \tgn \cite{tgn}, a memory-based \tgnn on real-world temporal interaction datasets, and establish that it consistently outperforms state-of-art methods on future link predictions task on target data. Next, we perform the ablation of \tgn model transfer with and without memory, showing that memory transfer plays a crucial role in target graph performance.  { Moreover, to strengthen this claim, we evaluate our proposed memory transfer techniques with non-memory based \tgnn \tgat \cite{tgat} and the recent state-of-art \gm \cite{gm}.}
\end{itemize}

\vspace{-0.15in}
\section{Preliminaries}

\subsection{Definitions}

\begin{defn}[Static Graph]
     A static graph $\CG = (\CV,\CE,\CX)$ consists of nodes $\CV = \{v_1,v_2,...,v_N\}$ where {$\vert\CV\vert$}=N and edges $\CE \in \CV \times \CV$ where {$\vert\CE\vert$}=M. $\CX$ is a set of feature vectors $\{\x_1,\x_2 \ldots \x_N\}$, where $\x_v \in \mathcal{R}^{\vert F_v\vert}$, $F_v$ is a set of features applicable for node $v$. Similarly each edge $(u,v)$ might have a feature/attribute vector denoted by $\x_{uv} \in \mathcal{R}^{\vert F_{uv}\vert}$ 
     \label{def:static_graph}
\end{defn}
\vspace{-0.15in}
\begin{defn}[Temporal Interaction Graph]
    A temporal interaction graph is defined as a timestamped sequence of events $\CG = \{e_1,e_2 \ldots e_N\}$. { Each event $e_i=(u,v,t,\x_u,\x_v,\x_{uv}(t))$ is an interaction (temporal edge) between nodes $u$ and $v$ at time $t \in \mathbb{R}_{\geq 0}$ where $u,v \in \CV$, $\x_u\in \mathbb{R}^{\vert F_u\vert}$ and $\x_v \in \mathbb{R}^{\vert F_v\vert}$ are node features and $\x_{uv}(t)\in \mathbb{R}^{\vert F_{uv}(t)\vert}$ is interaction feature at time $t$. $F_u$ is a set of available features for node $u$}.
    \label{def:tig}
\end{defn}
\vspace{-0.15in}
\begin{defn}[Continuous Time Future Link Prediction]
    Given a temporal interaction graph $\CG$ till time $t$, node pairs $u$ and $v$, future time stamp $t'$, future link prediction is the task of predicting whether there will be an interaction (edge existence) between nodes $u$ and $v$ at time $t'$. This can be modeled as a binary classification task where a label of $1$ denotes a link and $0$ as no link.
\end{defn}

\vspace{-0.15in}
\subsection{Static Graph Representation Learning}
In this section, we briefly go over the standard representation learning framework for static graphs. Given a static graph $\CG = (\CV,\CE,\CX)$, a typical Graph Neural Network (\gnn) follows the following framework to learn an embedding $\ch_u$ for each node $u \in \CV$.
Node embeddings are initialized using their attribute vectors at layer $0$,
\begin{equation}
    \ch_u^0 = \x_u \forall u \in \CV
\end{equation}
For consecutive layers, following steps are executed.\\
\noindent $\bullet$ \textbf{Step-1:} Computing messages from each node in the neighborhood
\begin{equation}
    \cmsg_u^l(v) = \text{\textsc{MESSAGE}}^l(\ch_u^{l-1},\ch_v^{l-1},\x_{uv})  \;\;\forall v \in \mathcal{N}_u 
\end{equation}
$\bullet$ \textbf{Step-2:} Aggregating these messages at each node $u \in \CV$
\begin{equation}
    \overline{\cmsg}_u^l(v) = \text{\textsc{AGGREGATE}}^l(\{\!\!\{\cmsg_u^l(v)\} \!\!\})  \;\;\forall v \in \mathcal{N}_u 
\end{equation}
$\bullet$ \textbf{Step-3:} Combining aggregated information with previous layer embedding
\vspace{-0.1in}
\begin{equation}
    \ch_u^l = \text{\textsc{COMBINE}}^l(\ch_u^{l-1},\overline{\cmsg}_u^l)
\end{equation}
Where $\text{\textsc{MESSAGE}}^l,\text{\textsc{AGGREGATE}}^l,\text{\textsc{COMBINE}}^l\;\; \forall l \in [1\ldots L]$ are neural network function including \textit{identity}, \textit{max-pool}, \textit{min-pool}, \textit{sum} and \textit{average}. $\CN_u$ is a 1-hop neighborhood of node $u$ and defined as  $\CN_u = \{v:(u,v)\in \CE\}$. These steps are repeated for $L$ layers to compute the final node representation to execute downstream tasks, e.g., node classification and link prediction. 
\vspace{-0.12in}
\subsection{Temporal Interaction Graph Representation Learning} 
In temporal interaction graphs, node embeddings must capture the dynamic behavior. Existing \tgnn computes time-dependent node embeddings at time $t$ by only attending to nodes interacted before time $t$.Formally, a temporal neighborhood of node $u$ in temporal interaction graph $\CG$ at time $t$ is defined as $\CN_u(t) = \{v:(u,v,t')\in \CE \;\vert \; t' < t\}$. Now, the following steps are executed to compute the dynamic node representations for a node $u \in \CV$ at time $t$.
    

\noindent $\bullet$ \textbf{Initialization:} $\ch_u^0(t) = \x_u + \cm_u(t)$ where $\cm_u(t)$ is memory of node $u$ at time $t$, updated whenever there was a new interaction $(u,v,t',\x_{uv}(t'))$  {at time $t' < t$} using a recurrent neural network.
\vspace{-0.05in}
\begin{equation}
\vspace{-0.155in}
        \overline{\cm}_u = \text{\textsc{MSG}}(\cm_u(t^-),\cm_v(t^-),\Delta t,\x_{uv}(t'))
        \label{eq:memory_compute}
\end{equation}
\vspace{-0.05in}
\begin{equation}
    \cm_u(t) = \text{\textsc{GRU}}(\overline{\cm}_u,\cm_u(t^-))
    \label{eq:memory_update}
\end{equation}
where $\Delta t$ is difference between $t'$ and time-stamp $t^-$  {which is last memory update time of node $u$}.  {$\text{\textsc{MSG}}$ is a $\text{\textsc{MLP}}$ based neural network. The memory of a node represents its past behavior in a compressed vector and facilitates improved temporal aggregation \cite{tgn}.  

Next, for every layer $l$, we skip \textbf{Step-1} and directly compute \textbf{Step-2} and \textbf{Step-3} as follows. 

\noindent $\bullet$ \textbf{Step - 2:} Aggregating the previous layer embedding from temporal neighbourhood $\CN_u(t)$ at each node $u \in \CV$ as follows.
\vspace{-0.06in}
\begin{equation}
    \overline{\ch}_u^l(v) = \text{\textsc{AGGREGATE}}^l(\{\!\!\{ (\ch_v^{l-1}(t),t-t_{uv},\x_{uv}) \} \!\!\})  \forall v \in \mathcal{N}_u(t)
    \label{eq:ttgn_aggregate}
\end{equation}
 {$t_{uv}$ latest interaction time between node $u$ and $v$ before time $t$.}
$\bullet$ \textbf{Step - 3:} Combining aggregated information with previous layer embedding as follows.
\vspace{-0.1in}
\begin{equation}
    \ch_u^l(t) = \text{\textsc{COMBINE}}^l(\ch_u^{l-1},\overline{\ch}_u^l)
    \label{eq:ttgn_combine}
\end{equation}

\text{\textsc{MSG}}, $\text{\textsc{AGGREGATE}}^l, \text{\textsc{COMBINE}}^l  \forall l \in [1\ldots L]$ are trainable neural networks.  {Training is enumerated via future link prediction loss. Eq. \ref{eq:memory_compute} and \ref{eq:memory_update} are executed with a batch of new interactions in the graph. Steps 2 \& 3 are repeated $L$ times to compute $L$ layered node representation.}
\vspace{-0.10in}
\section{Problem Formulation}
We now describe the following novel formulation.
\vspace{-0.05in}
\begin{prob}[Transfer learning framework for temporal interaction graphs]\hfill
\label{prob:1}

\noindent
\textbf{Input:} \textit{Given a rich temporal interaction source graph $\CG^{src}$ (Definition ~\ref{def:tig}), let $\tgnn_{\Theta^{src}}:\mathcal{V}\times \mathcal{V} \times \mathbb{R}_{\geq 0} \rightarrow \mathbb{R}_{\geq 0}$ be a temporal graph neural model parameterized by parameters $\Theta^{src}$ given as input which is trained on $\CG^{src}$  that maps a source node, target node and a time-stamp to  {positive} real number interpreted as the probability of link formation at time $t$. We are also given a target temporal interaction graph $\CG^{tgt}$ having insufficient interactions. Moreover, $\CG^{src}$ and $\CG^{tgt}$ are disjoint graphs, having no common nodes i.e. $\CV^{src} \cap \CV^{tgt}= \phi$. Also, the attribute vector dimension of the source and target graphs need not be the same, i.e., $\vert\x_{v^{src}}\vert \neq \vert\x_{v^{tgt}}\vert $.
\looseness=-1}

\noindent
\textbf{Goal:} \textit{Learn parameters $\Theta^{tgt}$ of temporal graph neural network on target graph  $\CG^{tgt}$ denoted as $\tgnn_{\Theta^{tgt}}$. $\Theta^{tgt}$ 
should leverage inductive biases contained in $\tgnn_{\Theta^{src}}$ without altering $\tgnn_{\Theta^{src}}$ with the following objectives.
\vspace{-0.05in}
\begin{itemize}
    \item \textbf{Testing Accuracy:} $\Theta^{tgt}$ should accurately predict the future link in chronologically split test interactions of the target graph.
    \item \textbf{Fast fine-tuning:} Optimum $\Theta^{tgt}$ should be achieved in few fine-tuning steps.
\end{itemize}}
\end{prob}
\vspace{-0.05in}
We focus on bipartite temporal interaction graphs where nodes are partitioned into users and items, $\CV^{src}=(\CU^{src},\CI^{src})$ and $\CV^{tgt}=(\CU^{tgt},\CI^{tgt})$. This simplification aids in experimentation and illustration, though this framework extends to non-bipartite interaction graphs as well.
 { Transferring \tgnn model requires transferring weights of neural modules in equations \ref{eq:memory_compute}, \ref{eq:memory_update},\ref{eq:ttgn_aggregate}  and \ref{eq:ttgn_combine}.To understand \textbf{why memory transfer is required}, let us assume max interaction time at $\CG^{src}$ as $T^{src}$. As transferred $\text{\textsc{GRU}}$'weights from $\CG^{src}$ to the target graph $\CG^{tgt}$ will be influenced by memory before time $T^{src}$ at $\CG^{src}$, replacing that memory with $\mathbf{0}$ at $\CG^{tgt}$ will result in deteriorated performance. Since memories are transductive, we formulate the following problem of memory transfer between $\CG^{src}$ and $\CG^{tgt}$ graph.}
\vspace{-0.1in}
\begin{prob}[Target domain transfer of source memory]\hfill
\label{prob:2}

\noindent
\textbf{Input:} \textit{A trained \tgnn parameters $\tgnn_{\Theta^{src}}$ on source $\CG^{src}$ graph with corresponding learned memory $\{\cm_u^{src} \;\forall u \in \CV^{src}\}$ and target data-scarce graph $\CG^{tgt}$
\looseness=-1}

\noindent
\textbf{Goal:} \textit{Learn a memory mapping function $f_{memory}: V(\CG^{src})\rightarrow V(\CG^{tgt})$ such that vertices in $\CG^{tgt}$ are mapped to vertices $\CG^{src}$  {having} most similar structural and behavior characteristics.}
\vspace{-0.1in}
\end{prob}

This mapping function need not be bijective, i.e., multiple nodes in  $\CG^{tgt}$ can map to the same node in $\CG^{src}$. Also we do not assume any restriction on a relative number of nodes between the source and target graph. 

\vspace{-0.1in}
\begin{figure}[t!]
    \centering
    \includegraphics[scale=0.33]{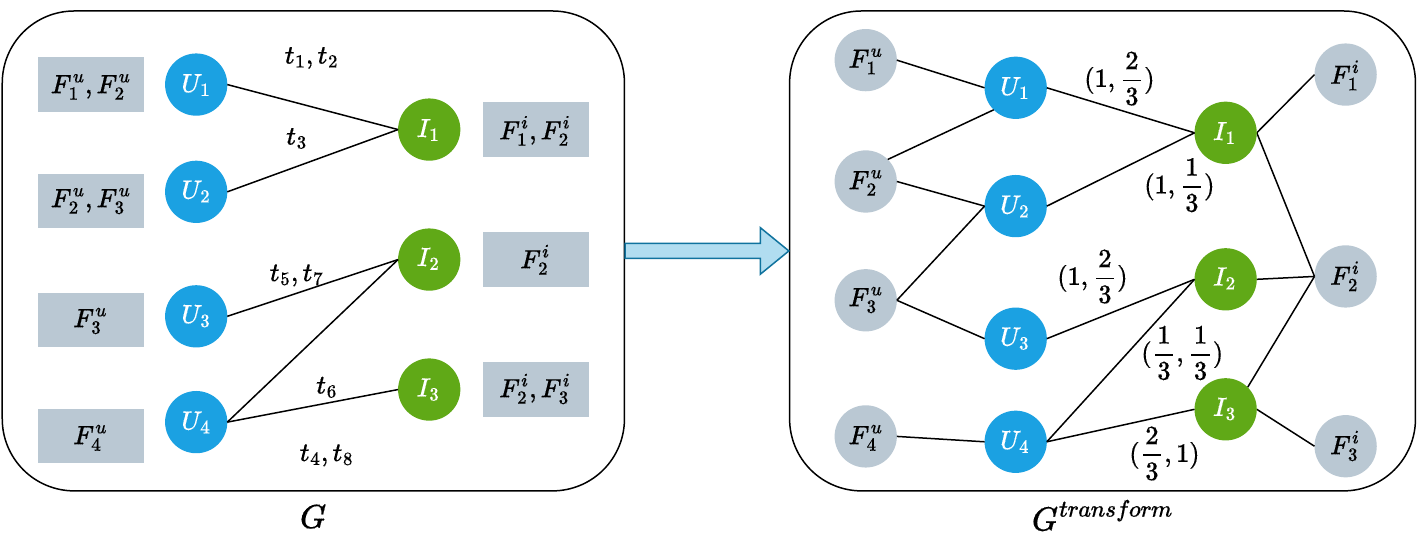}
    \vspace{-0.2in}
    \caption{Transformation of input graph $G$ to $G^{transform}$, where features are converted into nodes and repeated edges are replaced with their equivalent weight representation.}
    \label{fig:g_transform}
    \vspace{-0.2in}
\end{figure}
\section{\namemodelnew Description}
Our method consists of the following two steps:
\begin{itemize}
\item  \textbf{Step-1:} Learning a node mapping between source graph $\CG^{src}$ and target graph $\CG^{tgt}$.
\item \textbf{Step-2:}  Assigning memory to target graph nodes from corresponding source graph nodes, initializing equations \ref{eq:memory_update} and \ref{eq:memory_compute} with this memory, initializing weights of $\tgnn_{\Theta^{tgt}}$ with $\tgnn_{\Theta^{src}}$, finally fine-tune for few epochs on future link prediction loss on target data $\CG^{tgt}$
\end{itemize}

\subsection{Step-1: Mapping Memory from Source to Target}
 Problem \ref{prob:1} specifies two constraints:
 \begin{inparaenum}[(a)] \item  {The target graph has low interactions due to data scarcity, leading to no temporal patterns } \item Feature dimensions in both source and target graphs might be different. 
 \end{inparaenum} 
Since we wish to leverage common attributes between the source and target graph for mapping memory, we propose decoupling graph nodes from the attribute vector by creating virtual nodes where each virtual node corresponds to an attribute/feature. Specifically, we lose timestamp information for learning the memory mapping function and convert {both the source and target} temporal graph to a static one by assigning edge weights between user and item nodes based on interaction frequency.
\vspace{-0.1in}
\subsubsection{Graph Transformation}
Given a source temporal interaction graph  {$\CG=\{e_1,e_2 \ldots e_N\}$ \footnote{Here we interchangeably use $\CG^{src}$ with $\CG$}where each $e_i$ is an interaction between source $u$ and target node $v$ at time $t$ as defined in definition \ref{def:tig}}, we construct a static version $\CG^{static}=(\CV,\CE^{static})$ where,
\vspace{-0.05in}
\begin{equation}
    \CE^{static} = \{(u,v,a_{uv}) \mid (u,v,t,\x_u,\x_v,\x_{uv}(t)) \in \CG \}
\end{equation}
\vspace{-0.05in}
\begin{equation}
a_{uv} = \frac{\text{\# of interactions between nodes u and v}}{\text{Total \# of interactions by node $u$}}
\end{equation}
$\CG^{static}$ is a directed graph as $a_{uv}$ might not be the same as $a_{vu}$.
Given this graph $\CG^{static}=(\CV,\CE^{static})$, we now define a global feature set $F = \{f \mid f \;\in \cup F_v \; \forall v \in \CV \}$. The global feature set $F$ is converted to virtual nodes and added to $\CG^{static}$. We term this transformed graph as $\CG^{transform} = (\CV^{transform},\CE^{transform})$ where $\CV^{transform} = \CV \; {{\cup}} \;F$ and $\CE^{transform} = \CE^{static} \; {\cup} \; \CE^{feature}$ with $\CE^{feature}=\{(u,f) \mid  \forall f \in F_u \;\forall \; u \in \CV\}$, and  $F_u$ is set of available feature at node $u$.


Figure \ref{fig:g_transform} depicts this transformation, where $F^u_1$ at $U_1$ implies that $F_1^u$ is available for user $U_1$. We denote $F^u$ for user feature nodes and $F^i$ for item feature nodes. User and item nodes are assigned $0$ vectors as feature vectors after this transformation. This transformation has the following benefits.
\begin{itemize}
    \item This graph transformation introduces feature nodes, which become common nodes between the transformed source graph and the target graph.
    \item  These common feature nodes will allow learning node representations based on common knowledge between source and target graphs derived from structural and behavior patterns. 
    \item Memory mapping is now feasible by mapping the target graph nodes to the source graph nodes via distance in embedding space generated by training a graph representation learning model on a transformed source graph.
    \item Representation learning model needs to be trained once, which is reusable across multiple target graphs.
\end{itemize}


\vspace{-0.1in}
\subsubsection{Representation learning on transformed graph}
\begin{figure*}[!ht]
\centering
  \vspace{-0.15in}
  \includegraphics[scale=0.30]{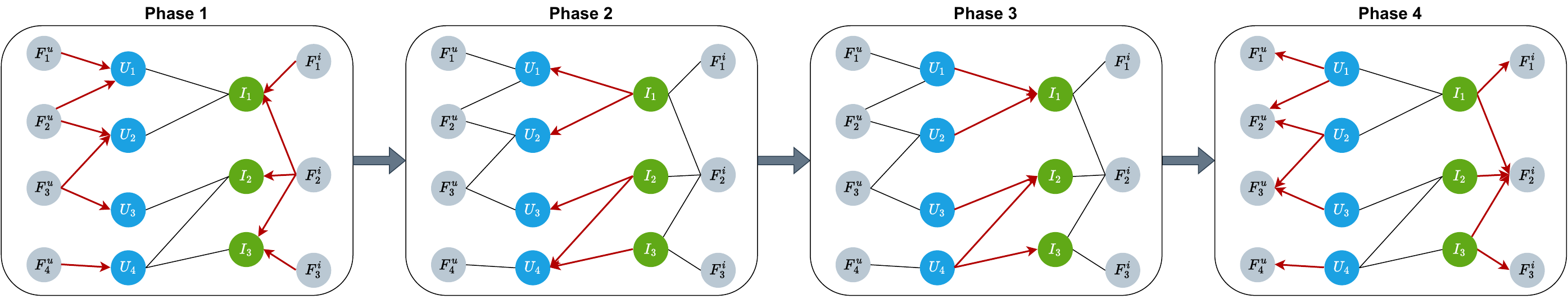}
  \vspace{-0.15in}
  \caption{ A breakdown of message passing pipeline architecture. Phase 1 corresponds to information passed from feature nodes to user and item nodes, Phase 2 corresponds to information passed from item nodes to user nodes, Phase 3 corresponds to information passed from user nodes to item nodes, and Phase 4 corresponds to information passed from user and item nodes back to feature nodes.}
  \label{fig:step-wise-msg-passing}
  \vspace{-0.1in}
\end{figure*}
Each node in $\CG^{transform}$ has two type of neighborhoods: \textbf{a)} graph node \textbf{b)} feature node, defined as follows:

\noindent \textbf{Graph node neighborhood for graph node $u \in \CV$:} 
  \vspace{-0.05in}
\begin{equation}
    \mathcal{N}_u^{graph} = \{v \mid (u,v) \in \CE^{static} \}
    \label{eq:graph_node_nbr}
\end{equation}
\noindent \textbf{Graph node neighborhood for features node  $f \in F$:} 
  \vspace{-0.05in}
\begin{equation}
    \mathcal{N}_f^{graph} = \{v \mid (v,f) \in \CE^{feature} \}
    \label{eq:graph_node_nbr_feature_node}
\end{equation}
\noindent \textbf{Feature node neighborhood for graph node $u \in \CV$:}
  \vspace{-0.05in}
\begin{equation}
    \mathcal{N}_u^{feature} =  \{f \mid (u,f) \in \CE^{feature} \}
    \label{eq:feature_node_nbr}
\end{equation}
We now propose a node learning representation learning mechanism with these neighborhood definitions, incorporating feature nodes on a transformed source graph $\CG^{transform}$. The goal is to generate representation on any target graph once the model is trained. To achieve this, we define the following embedding initialization on $\CG^{transform}$
\vspace{-0.05in}
\begin{equation}
\ch_v^0 = \mathbf{0} \; \; \forall v \in \CV
\label{eq:gat_init}
\end{equation}
\vspace{-0.2in}
 {\begin{equation}
\ch_f^0 = \text{\textsc{Embedding}}(f) \;\; \forall f \in F
\end{equation}}
where \text{\textsc{Embedding}} is a trainable layer. Since graph nodes have initial embedding set to $\mathbf{0}$, a standard \gnn cannot capture topological information at the next layer. To address this, we propose a novel 4-stage message-passing framework on $\CG^{transform}$ to learn the representation for both graph and feature nodes. 
We first define an aggregation function $\text{{g}}_{\theta}$ that computes the target node's representation based on its either graph or feature node neighbors. Specifically, 
\begin{equation}
    \ch_u^l = \text{{g}}_{\theta}(\ch_u, \mathcal{N}_u^{\text{custom}},l,l_{\mathcal{N}_u^{\text{custom}}})
\end{equation}
where $\ch_u$ and $\ch_u^l $ represents the input and updated embedding of node $u$ respectively, $l$ is current layer. The neighborhood $\mathcal{N}_u^{\text{custom}}$ can either be graph-node-based or feature-node-based. Additionally, $l_{\mathcal{N}_u^{\text{custom}}}$ is the layer number of neighbors embeddings to be used in $\text{{g}}_{\theta}$, its application will be clearer in the steps below. The specific semantic of $\text{{g}}_{\theta}$ is as follows.
\vspace{-0.1in}
\begin{alignat}{3}
\nonumber
    \cmsg_u^{l}(v)&{{=}\text{\textsc{LeakyReLU}}\left({\cw_1^l}\ch_u{ \parallel}{\cw_2^{l}}\ch_v^{l_{\mathcal{N}_u^{\text{custom}}}}{ \parallel}w^{l}_3a_{uv}\right)}\; \forall v \in  \mathcal{N}_u^{\text{custom}} \\
 \nonumber
    \alpha_{vu}&{ =} {\frac{\exp\left({\cw_{4}^{l}}^T\cmsg^{l}_u\left(v\right)\right)}{\sum\limits_{v' \in \mathcal{N}_{u}^{feature}}\exp\left({\cw^{l}_{4}}^T\cmsg^{l}_u\left(v'\right)\right)}} \; \forall v \in  \mathcal{N}_u^{\text{custom}} \\
   { \ch_u^{l}}&{ = }\text{\textsc{MLP}}_1^l\left(\cw^{l}_{5}\ch_u \parallel {\sum_{v \in \mathcal{N}_{u}^{feature}}}{ \alpha_{vu}\cw^{l}_{6}\ch_v^{l_{\mathcal{N}_u^{\text{custom}}}}} \right)
\label{eq:F_theta}
\end{alignat} 

Given this definition, each layer $l$ comprises the following phases.

\noindent \textbf{Phase 1:} Information is exchanged from feature nodes to user and item nodes, followed by aggregation at graph nodes (user and item) and combining this new information with previous layer embedding. We adopt $\text{g}_{\theta}$ based aggregation to compute the user and item node embeddings. $\forall u \in \CV$
\vspace{-0.05in}
\begin{equation}
    \ch_u^{l}=\text{{g}}_{\theta_1^l}(\ch_u^{l-1},\mathcal{N}_u^{feature},l,l-1)
    \label{eq:step1_agg}
\end{equation}

\noindent \textbf{Phase 2:} Now we exchange information from item nodes to user nodes and repeat the same steps of aggregation and combining. $\forall u \in \CU$
\vspace{-0.05in}
\begin{equation}
    \ch_u^{l}=\text{{g}}_{\theta^l_2}(\ch_u^l,\mathcal{N}_u^{graph},l,l)
    \label{eq:step2_aggregation}
\end{equation}

\noindent \textbf{Phase 3:} This information is now exchanged from user to item nodes, similar to phase 2.$\forall u \in \CI$
\vspace{-0.05in}
\begin{equation}
    \ch_u^{l}=\text{{g}}_{\theta^l_3}(\ch_u^l,\mathcal{N}_u^{graph},l,l)
    \label{eq:step3_aggregation}
\end{equation}

\noindent \textbf{Phase 4:} Finally, the information at the user and item nodes is sent back to feature nodes. $\forall u \in F$
\vspace{-0.05in}
\begin{equation}
    \ch_u^{l}=\text{{g}}_{\theta^l_4}(\ch_u^l,\mathcal{N}_u^{graph},l,l)
    \label{eq:step4_aggregation}
\end{equation}

\begin{figure*}[h]
    \vspace{-0.2in}
    \centering
    \includegraphics[scale=0.3]{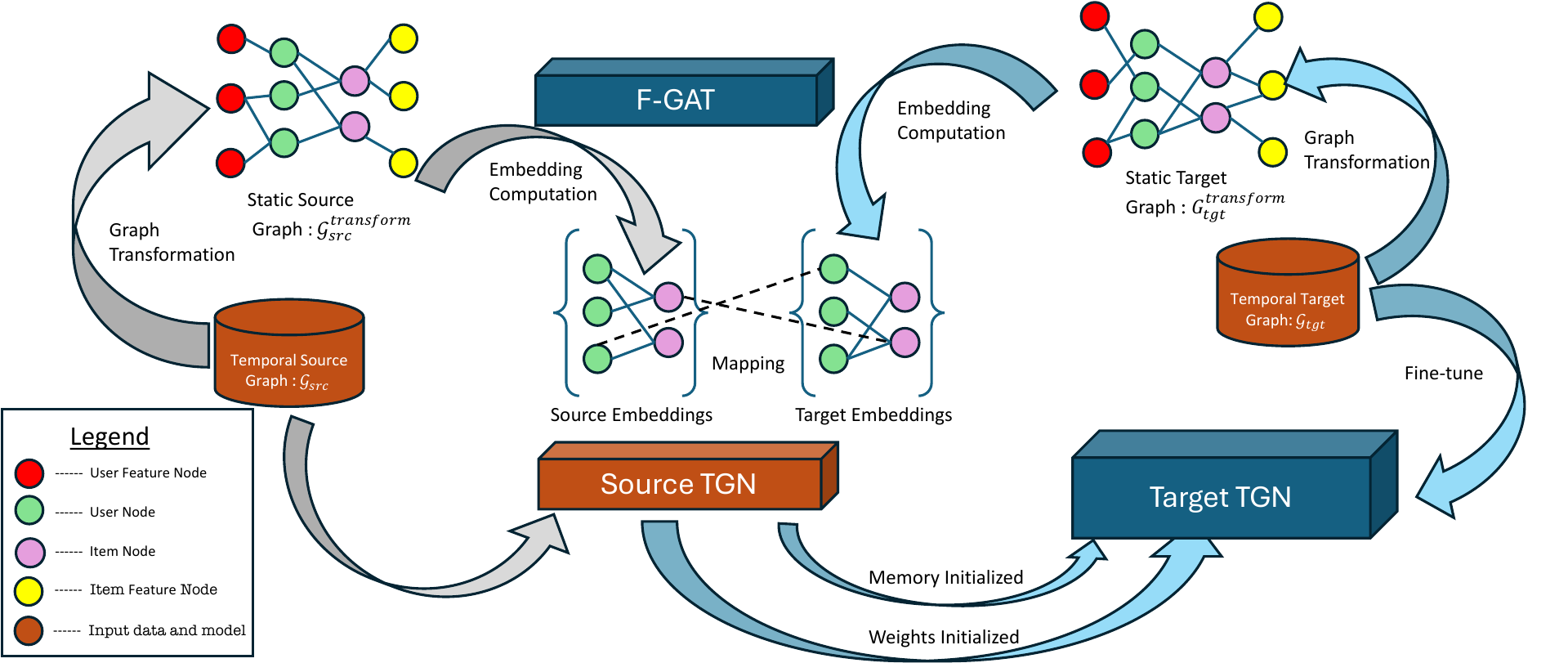}
    \vspace{-0.1in}
    \caption{Overall architecture of \namemodelnew framework. First, the source temporal graph $G_S$ and target temporal graph $G_T$ are converted into their respective static transformation graphs. These are then passed to \fgat to compute their node embeddings. These node embeddings are then used to create a source-target node memory mapping. Now, weights and mapped memory are transferred from trained \tgn on the source graph to initialize \tgn on the target graph. Finally, target \tgn is fine-tuned on the target graph}
    \vspace{-0.18in}
    \label{fig:mint_architecture}

\end{figure*}

These steps are repeated $L$ times to generate a rich representation for user and item nodes.  $\text{{g}}_{\theta_i}^l \;\forall i \in \{1..4\} \; \& \;l \in \{1..L\}$  are trainable neural functions.  {We note that phase 2 and phase 3 are interchangeable as information flow can be from user to item or item to user. Further, in a non-bipartite setting, phases 2 and 3 will be merged into a single phase.  The order of phases 1 and 4 should remain as defined since eq. \ref{eq:gat_init} assigns $\mathbf{0}$ vector to graph nodes. Thus, they require information first from feature nodes to enable phases 2 and 3.} These steps are also described in figure \ref{fig:step-wise-msg-passing}. We term this \gnn as \textit{Feature Graph Attention Network} (\fgat).

\vspace{-0.10in}
\subsubsection{Memory Transfer}
Our proposed transformation is generic and facilitates memory transfer from one source graph to another graph. To achieve this, we first train \fgat using a link-prediction task on a collection of graphs similar to source graph, including source graph, even if these graphs are not temporal, by applying the same transformation.  These diverse set of training graphs facilitates robust learning of \fgat. After training, we obtain node representation on both $G^{src}$ and $G^{tgt}$ using \fgat.  
\vspace{-0.1in}
\begin{equation}
    \cH^{src} = \fgat(\CG^{src^{transform}})
\end{equation}
\vspace{-0.2in}
\begin{equation}
    \cH^{tgt} = \fgat(\CG^{tgt^{transform}})
\end{equation}
where $\cH^{src} = {\ch_u \;\forall\; u \in \CV^{src}}$ and $\cH^{tgt} = {\ch_u \;\forall \; u \in \CV^{tgt}}$.

Finally, we map the memory of target nodes from source graph nodes as follows, $\forall u \in \CV^{tgt}$
\vspace{-0.1in}
\begin{equation}
    src\_node = \arg\max_{v \in \CV^{src}}(\ch^{src^T}_v\ch^{tgt}_u)
\end{equation}
\vspace{-0.2in}
\begin{equation}
\cm_u^{tgt} = \cm^{src}_{src\_node}
\end{equation}
where $\cm_{src\_node}^{src}$ is the input memory of the source graph as defined in problem statement \ref{prob:2}. We find this simple initialization method effective, though more complex mapping functions can be designed to utilize a combination of the top closest source nodes to initialize the target node's memory.

\vspace{-0.10in}
\subsubsection{Transfer Learning on $\CG^{tgt}$}
After initializing the memory of the target graph, we initialize the weights  {$\Theta^{tgt}$} of $\tgnn_{\Theta^{tgt}}$ with weights $\Theta^{src}$ of trained source \tgnn. Finally, the target model is trained on the target training graph for a few epochs on future link prediction tasks as standard practice in few-shot fine-tuning. 

Figure \ref{fig:mint_architecture} summarizes the proposed \namemodelnew framework on temporal interaction graphs.



\vspace{-0.15in}
\section{Experiments}
In this section, we evaluate our proposed framework \namemodelnew on transfer capabilities and answer the following questions?

\noindent\textbf{RQ1:} Can \namemodelnew outperform no-transfer \tgnn and state-of-the-art baselines on data-scarcity conditions?

\noindent\textbf{RQ2:} Is memory-transfer module of \namemodelnew significant compares to transferring only weights?

\noindent\textbf{RQ3:} Is \namemodelnew required even without low-data conditions in target graphs?

\vspace{-0.1in}
\subsection{Dataset}

To simulate our transfer conditions as outlined in problem \ref{prob:1}, we work with a popular check-in dataset by Foursquare \cite{84766bf19fce41aa8d3051004dd468b2}, which includes long-term (about 22 months from Apr. 2012 to Jan. 2014) global-scale check-in data collected from over $400$ cities. It is a temporal bipartite interaction graph between users and businesses for all $400$ cities. We refer to businesses as items in the proposed methodology. We evaluate the transferring capabilities of \namemodelnew over check-in recommendations as future link prediction tasks from a particular source graph dataset to various target graphs. We filter out businesses with less than 20 check-ins and users with less than 10 check-ins for source datasets. These thresholds are set for target datasets at 10 and 2, respectively. A higher threshold is used for source datasets to minimize the effects of noisy data during transfer. Tables \ref{tab:combined_city_stats} presents the statistics for different source and target cities, highlighting the significantly low interactions in target cities compared to source cities. Additionally, we evaluate the efficacy of \namemodelnew on Wiki \cite{wiki_dataset} and Yelp dataset \cite{yelp_dataset}.



 



\begin{table}[h]
\vspace{-0.15in}
\caption{Number of interactions in training data of Foursquare datasets}
\centering
\vspace{-0.15in}
\resizebox{0.35\textwidth}{!}{%
\begin{tabular}{llll}
\toprule
\textbf{Source City} & \textbf{Interactions} & \textbf{Target City} & \textbf{Interactions} \\
\midrule
Moscow        & 525520 & Brussels        & 3934  \\
Jakarta       & 538262 & Campo Grande    & 5719  \\
Tokyo         & 562962 & Rio de Janeiro  & 3534  \\
Kuala Lumpur  & 533927 & Gent            & 2961  \\
              &        & San Antonio     & 4210  \\
              &        & Berlin          & 3336  \\
\bottomrule
\end{tabular}
}
\label{tab:combined_city_stats}
\end{table}

\begin{table*}[!ht]
\vspace{-0.1in}
\caption{Transfer comparison on various pairs from Foursquare}
\vspace{-0.1in}
\centering
\resizebox{0.6\textwidth}{!}{%
\begin{tabular}{llcccc}


\toprule
\textbf{Transfer} & \textbf{Method} & \textbf{AP} &\textbf{AUC} &  \textbf{MRR} & \textbf{Recall}\boldmath{$@20$}\\
\midrule
 \multirow{5}{*}{MS $\rightarrow$ GN} & NGCF & $ 0.6892 \pm 0.0 $& $0.6471 \pm 0.0$& $0.1156 \pm 0.0$& $0.2031 \pm 0.0$ 
 \\
 & LightGCN & $0.6911 \pm 0.0$& $0.6529 \pm 0.0$& $0.1220 \pm 0.0 $& $0.2110 \pm 0.0$ \\

 \cmidrule{2-6}
& NT-TGN &$ 0.6035 \pm  0.0027 $&$ 0.6086 \pm  0.0062 $&$ 0.0225 \pm 0.0012 $&$ 0.0684 \pm 0.014 $\\
& WT-TGN &$ 0.8182 \pm 0.0143 $&$ 0.8232 \pm 0.0113 $&$ 0.1256 \pm 0.0139 $&$ 0.3136 \pm 0.0237 $\\
& \mtgn &$ \mathbf{0.9426 \pm 0.0015} $&$ \mathbf{0.937 \pm 0.002} $&$ \mathbf{0.332 \pm 0.0069} $&$ \mathbf{0.6598 \pm 0.0048} $\\

\cmidrule{1-6}

 \multirow{5}{*}{MS $\rightarrow$ BR} & NGCF & $0.7129 \pm 0.0$& $0.6736 \pm 0.0$& $0.1259 \pm 0.0$& $0.2399 \pm 0.0$ \\

 & LightGCN & $0.7213 \pm 0.0$& $0.6802 \pm 0.0$& $0.1289 \pm 0.0$& $0.2439 \pm 0.0$ \\

 \cmidrule{2-6}
& NT-TGN &$ 0.6488 \pm 0.0024 $&$ 0.6674 \pm  0.0081 $&$ 0.0183 \pm 0.0061 $&$ 0.062 \pm 0.0091 $\\
& WT-TGN &$ 0.8649 \pm 0.0084 $&$ 0.8678 \pm 0.0081 $&$ 0.1385 \pm 0.0073 $&$ 0.3876 \pm 0.0156 $\\
& \mtgn &$ \mathbf{0.9407 \pm 0.0028} $&$ \mathbf{0.9346 \pm 0.0031} $&$ \mathbf{0.3335 \pm 0.0134} $&$ \mathbf{0.6437 \pm 0.0067} $\\

 \cmidrule{1-6}

  \multirow{5}{*}{JK $\rightarrow$ KW} & NGCF & $ 0.605 \pm 0.0$& $0.5425 \pm 0.0$& $0.0840 \pm 0.0$& $0.1527 \pm 0.0$ \\

 & LightGCN & $0.6467 \pm 0.0$& $ 0.6075\pm 0.0$& $0.0898 \pm 0.0$& $0.1681 \pm 0.0$ \\

 \cmidrule{2-6}
& NT-TGN &$ 0.7245 \pm 0.0073 $&$ 0.7313 \pm 0.0066  $&$ 0.0502 \pm 0.0158 $&$ 0.197 \pm 0.0165 $\\
& WT-TGN &$ 0.7982 \pm 0.0029 $&$ 0.8123 \pm 0.0 $&$ 0.0963 \pm 0.0091 $&$ 0.3152 \pm 0.0185 $\\
& \mtgn &$ \mathbf{0.9072 \pm 0.0} $&$ \mathbf{0.8996 \pm 0.0011} $&$ \mathbf{0.3126 \pm 0.0023} $&$ \mathbf{0.5607 \pm 0.0013} $\\

  \cmidrule{1-6}

  \multirow{5}{*}{JK $\rightarrow$ RJ} & NGCF & $0.6234 \pm  0.0037$& $0.588 \pm 0.0029$& $0.0714 \pm  0.0012$& $0.1225 \pm 0.0021$ \\
  
 & LightGCN & $0.6288 \pm 0.0$& $0.5947 \pm 0.0$& $0.0760 \pm 0.0$& $0.1281 \pm 0.0$ \\

 \cmidrule{2-6}
& NT-TGN &$ 0.5974 \pm 0.0035 $&$ 0.6151 \pm 0.0061 $&$ 0.0159 \pm 0.0052 $&$ 0.0472 \pm 0.0129 $\\
& WT-TGN &$ 0.888 \pm 0.0078 $&$ 0.8914 \pm 0.0077 $&$ 0.1425 \pm 0.0099 $&$ 0.3903 \pm 0.014 $\\
& \mtgn &$ \mathbf{0.9478 \pm 0.0038} $&$ \mathbf{0.9415 \pm 0.0039} $&$ \mathbf{0.3504 \pm 0.017} $&$ \mathbf{0.6689 \pm 0.0092} $\\

   \cmidrule{1-6}

 \multirow{5}{*}{TY $\rightarrow$ CG} & NGCF & $0.6472 \pm 0.0$& $0.6185 \pm 0.0$& $0.06536 \pm 0.0$& $0.1379 \pm 0.0$ \\
 
 & LightGCN & $0.6635 \pm 0.0$& $0.6267 \pm 0.0$& $0.0715 \pm 0.0$& $0.1523 \pm 0.0$ \\

 \cmidrule{2-6}
& NT-TGN &$ 0.6197 \pm 0.0091 $&$ 0.6331 \pm 0.0154 $&$ 0.0157 \pm 0.0032 $&$ 0.0545 \pm 0.0085 $\\
& WT-TGN &$ 0.8633 \pm 0.0058 $&$ 0.8502 \pm 0.0062 $&$ 0.1207 \pm 0.0027 $&$ 0.3455 \pm 0.0041 $\\
& \mtgn &$ \mathbf{0.9118 \pm 0.0} $&$ \mathbf{0.9027 \pm 0.002} $& $\mathbf{0.2407 \pm 0.0012} $&$ \mathbf{0.5343 \pm 0.0022} $\\

\cmidrule{1-6}

 \multirow{5}{*}{TY $\rightarrow$ SA} & NGCF & $ 0.6239 \pm 0.0$& $0.5734 \pm 0.0$& $0.0703 \pm 0.0$& $0.1607 \pm 0.0$ \\
 & LightGCN & $0.6719 \pm 0.0104$& $0.632 \pm 0.0065$& $0.08418 \pm 0.0087$& $0.1788 \pm 0.0039$ \\

 \cmidrule{2-6}
& NT-TGN &$ 0.5674 \pm 0.0035 $&$ 0.565 \pm 0.0019 $&$ 0.0097 \pm 0.0043 $&$ 0.0352 \pm 0.0108 $\\
& WT-TGN &$ 0.8397 \pm 0.0024 $&$ 0.8187 \pm 0.0029 $&$ 0.1157 \pm 0.0049 $&$ 0.3091 \pm 0.008 $\\
& \mtgn &$ \mathbf{0.877 \pm 0.0014} $&$ \mathbf{0.8577 \pm 0.0012} $&$ \mathbf{0.2335 \pm 0.0034} $&$ \mathbf{0.455 \pm 0.0037} $\\

\bottomrule
\vspace{-0.15in}
\end{tabular}%
\vspace{-0.2in}
}
\label{tab:main_results5}
\end{table*}
\vspace{-0.2in}
\subsection{Experimental Settings}
This section describes the experimental settings employed in our research study. \textbf{Transfer Protocol:} For each source region dataset in Foursquare, we employ a temporal data split, wherein the dataset 
was partitioned into training (70\%), validation (15\%), and test (15\%) sets  {sequentially on sorted time-stamped interactions}. We train our source models for 10 epochs each. A split of 10\% for training and 45-45\% for validation and testing sets was adopted for each target region dataset to simulate data scarcity. We deliberately opted for a 10\%  allocation for the training set in each target dataset, aiming to create a sparse training environment suited for evaluating the transferring capabilities of \namemodelnew. For source graphs in transfer, we select the top 4 graphs in terms of the number of interactions. For target graphs, we pick all the graphs with the total number of un-preprocessed interactions between 40,000 and 80,000, giving us 79 target graphs. We note that since we use only 10\% data for training on the target graph, the \# of interactions in target training data varies from $2000$ to $8000$ \ref{tab:combined_city_stats} after pre-processing. This also enables us to have sufficient \# of interactions($20-40$K) in test target data to evaluate the quality of the trained model robustly. To train \fgat, we created a pool of the top 30 graphs similar to the source graph. We don't include any target graph in this pool. We temporally partition the data of each of the 30 graphs into a train/validation/test of 70\%-15\%-15\%. We randomly sample one of these graphs at each epoch and train on masked link prediction loss. Specifically, we mask 50\% of training edges, only using the remaining 50\% for message passing and back-propagating loss computed using masked 50\% edges. We employ this training strategy to ensure that our \fgat learns from a broad range of data distributions and is less susceptible to over-fitting on a particular graph. This \fgat, once trained, can generate source-target graph memory mapping without any further training for any number of target datasets. We evaluate all approaches over 5 independent runs for the transfer setting and report the mean and standard deviation.  {A similar strategy is followed for Wiki and Yelp datasets. We use the same \tgnn hyper-parameters reported in menu-script \cite{tgn}. All our models are implemented in PyTorch on a server running Ubuntu 18.04.4 LTS. CPU: Intel(R) Xeon(R) Gold 6248 CPU @ 2.50GHz, RAM: 385GB, and GPU: NVIDIA V100 GPU.  


\begin{table}[h]
\caption{Comparison with non-memory based \tgnn, \tgat and \gm on Foursquare}
\vspace{-0.1in}
\centering
\resizebox{1.0\columnwidth}{!}{%
\begin{tabular}{llccccc}
\toprule
\multirow{ 2}{*}{\textbf{Transfer}} & \multirow{2}{*}{\textbf{Metric}} & \multicolumn{2}{c}{\textbf{\tgat}} &  \multicolumn{2}{c}{\textbf{\gm}} & \multirow{ 2}{*}{\textbf{MINTT}}\\
\cmidrule{3-4}\cmidrule{5-6}
& & \textbf{NT} & \textbf{WT} & \textbf{NT} & \textbf{WT} & \\
\midrule
\multirow{3}{*}{MS $\rightarrow$ GN}  & AP & $0.7285 \pm 0.009$ & $0.8720 \pm 0.007$ & $0.7442 \pm 0.005$ & $0.8746 \pm 0.000$ & $\mathbf{0.9426 \pm 0.002}$ \\
& AUC & $0.7216 \pm 0.002$ & $0.8649 \pm 0.008$ &  $0.7161 \pm 0.008$ & $0.8698 \pm 0.000$ & $\mathbf{0.9370 \pm 0.002}$\\
& MRR & $0.0379 \pm 0.009$ & $0.1857 \pm 0.004$ & $0.0792 \pm 0.003$ & $0.1459 \pm 0.002$ & $\mathbf{0.3320 \pm 0.007}$\\
\cmidrule{1-7}
\multirow{3}{*}{MS $\rightarrow$ BR} 
& AP & $0.7347 \pm 0.013$ & $0.8828 \pm 0.002$ & $0.7411 \pm 0.004$ & $0.8723 \pm 0.001$ & $\mathbf{0.9407 \pm 0.003}$\\
& AUC & $0.7184 \pm 0.007$ & $0.8751 \pm 0.002$ & $0.7100 \pm 0.007$ & $0.8680 \pm 0.002$ & $\mathbf{0.9346 \pm 0.003}$\\
& MRR & $0.0236 \pm 0.008$ & $0.1765 \pm 0.004$ & $0.0655 \pm 0.007$ & $0.1285 \pm 0.000$ & $\mathbf{0.3335 \pm 0.013}$\\
\cmidrule{1-7}
\multirow{3}{*}{JK $\rightarrow$ KW} 
& AP & $0.7928 \pm 0.014$ & $0.8767 \pm 0.010$ &  $0.8044 \pm 0.007$ & $0.8901 \pm 0.003$ & $\mathbf{0.9072 \pm 0.000}$\\
& AUC & $0.7719 \pm 0.017$ & $0.8742 \pm 0.011$ & $0.7800 \pm 0.011$ & $0.8860 \pm 0.000$ & $\mathbf{0.8990 \pm 0.001}$\\
& MRR & $0.0612 \pm 0.009$ & $0.2031 \pm 0.007$ & $0.1238 \pm 0.003$ & $0.1724 \pm 0.001$ & $\mathbf{0.3126 \pm 0.002}$\\
\cmidrule{1-7}
\multirow{3}{*}{JK $\rightarrow$ RJ} 
& AP & $0.7458 \pm 0.012$ & $0.8892 \pm 0.002$ & $0.7428 \pm 0.007$ & $0.8865 \pm 0.000$ & $\mathbf{0.9478 \pm 0.004}$\\
& AUC & $0.7545 \pm 0.011$ & $0.8845 \pm 0.001$ & $0.7214 \pm 0.011$ & $0.8862 \pm 0.000$ & $\mathbf{0.9415 \pm 0.004}$\\
& MRR & $0.0310 \pm 0.002$ & $0.2060 \pm 0.007$ & $0.0495 \pm 0.000$ & $0.1200 \pm 0.009$ & $\mathbf{0.3504 \pm 0.017}$\\
\cmidrule{1-7}
\multirow{3}{*}{TY $\rightarrow$ CG} 
& AP & $0.7261 \pm 0.020$ & $0.8632 \pm 0.007$ & $0.7455 \pm 0.013$ & $0.8699 \pm 0.000$ & $\mathbf{0.9118 \pm 0.000}$\\
& AUC & $0.7058 \pm 0.023$ & $0.8527 \pm 0.009$ & $0.7201 \pm 0.021$ & $0.8656 \pm 0.001$ & $\mathbf{0.9027 \pm 0.002}$\\
& MRR & $0.0407 \pm 0.006$ & $0.0970 \pm 0.004$ & $0.0445 \pm 0.007$ & $0.0820 \pm 0.004$ & $\mathbf{0.2407 \pm 0.001}$\\
\cmidrule{1-7}
\multirow{3}{*}{TY $\rightarrow$ SA} 
& AP & $0.6818 \pm 0.047$ & $0.8473 \pm 0.008$ &  $0.7204 \pm 0.007$ & $0.8453 \pm 0.000$ & $\mathbf{0.8770 \pm 0.001}$\\
& AUC & $0.6778 \pm 0.026$ & $0.8352 \pm 0.011$ &  $0.6939 \pm 0.011$ & $0.8374 \pm 0.000$ & $\mathbf{0.8577 \pm 0.001}$\\
& MRR & $0.0158 \pm 0.000$ & $0.1336 \pm 0.003$ & $0.0435 \pm 0.008$ & $0.0975 \pm 0.003$ & $\mathbf{0.2335 \pm 0.003}$\\
\bottomrule
\end{tabular}

}
\label{tab:main_results_tgat_gm}
\vspace{-0.2in}
\end{table}

\begin{table}[h]
\caption{Average performance across all 79 target graphs for given sources from Foursquare}
\vspace{-0.1in}
\centering
\resizebox{0.9\columnwidth}{!}{%
\begin{tabular}{llccc}


\toprule
\textbf{Source} & \textbf{Method} & \textbf{AP} &\textbf{AUC} &  \textbf{MRR}\\
\midrule
\multirow{3}{*}{MS} 
& NT-TGN &$ 0.6461 \pm 0.0 $&$ 0.6485 \pm 0.0 $&$ 0.028 \pm 0.0 $\\
& WT-TGN &$ 0.8072 \pm 0.0073 $&$ 0.8073 \pm 0.0074 $&$ 0.1083 \pm 0.0056 $\\
& \mtgn &$ \mathbf{0.8746 \pm 0.0071} $&$ \mathbf{0.8673 \pm 0.007} $&$ \mathbf{0.2051 \pm 0.0098} $\\

\cmidrule{1-5}
\multirow{3}{*}{JK} 
& NT-TGN &$ 0.6461 \pm 0.0 $&$ 0.6485 \pm 0.0 $&$ 0.028 \pm 0.0 $\\
& WT-TGN &$ 0.8141 \pm 0.011 $&$ 0.8151 \pm 0.0099 $&$ 0.1052 \pm 0.0079 $\\
& \mtgn &$ \mathbf{0.87 \pm 0.0071} $&$ \mathbf{0.8644 \pm 0.0069} $&$ \mathbf{0.186 \pm 0.0079} $\\

 \cmidrule{1-5}
\multirow{3}{*}{TY} 
& NT-TGN &$ 0.6461 \pm 0.0 $&$ 0.6485 \pm 0.0 $&$ 0.028 \pm 0.0 $\\
& WT-TGN &$ 0.8446 \pm 0.0048 $&$ 0.8299 \pm 0.0051 $&$ 0.1679 \pm 0.0065 $\\
& \mtgn &$ \mathbf{0.8727 \pm 0.0021} $&$ \mathbf{0.8641 \pm 0.0022} $&$ \mathbf{0.2065 \pm 0.003} $\\

  \cmidrule{1-5}

\multirow{3}{*}{KL} 
& NT-TGN &$ 0.6461 \pm 0.0 $&$ 0.6485 \pm 0.0 $&$ 0.028 \pm 0.0 $\\
& WT-TGN &$ 0.8387 \pm 0.007 $&$ 0.8298 \pm 0.0069 $&$ 0.136 \pm 0.0069 $\\
& \mtgn &$ \mathbf{0.8641 \pm 0.0054} $&$ \mathbf{0.8584 \pm 0.0048} $&$ \mathbf{0.1744 \pm 0.0058} $\\

\bottomrule
\end{tabular}%

}
\label{tab:main_results_6}
\vspace{-0.2in}
\end{table}
\vspace{-0.1in}
\subsection{Baselines}

To compare our approach, we evaluate the performance of two popular state-of-the-art models for graph-based recommendations. We select the following baselines since they work well under data scarcity. \textbf{\lightgcn}\cite{lightgcn} simplifies the design of \ngcf with less \# of parameters by removing complex message and aggregation functions to make it more concise and appropriate for the link-prediction tasks, including only the most essential component in \gcn \cite{gcn} neighborhood aggregation for collaborative filtering. This is shown to work well in data-scarce settings. \textbf{\ngcf} \cite{NGCF} adopts the user-item bipartite graph to incorporate high-order relations using \gnn to enhance collaborative patterns. We use a 2-layer architecture for both \ngcf and \lightgcn. In \lightgcn, repeated interactions were accounted for, whereas \ngcf formulation only takes unique interactions. The train-val-test split is employed in the same way as described earlier. We train \lightgcn up to convergence and \ngcf for 25 epochs. Because of their transductive nature, we did not show transfer for \ngcf and \lightgcn.  
\vspace{-0.1in}
\subsubsection{Non-memory based \tgnn}: To showcase the importance of memory in model transfer for temporal interaction graphs, we also include non-memory based \tgnn \tgat\cite{tgat} and state-of-the-art model \gm\cite{gm} as baselines and compare with their weight transfer variant.}



\vspace{-0.2cm}
\subsubsection{Proposed Method Variants}
We propose three variants of the proposed method \mtgn and benchmark their performance in the transfer setting. \textbf{No-memory and no-weight transfer [NT-TGN]}- We train \tgn \cite{tgn} from scratch for 30 epochs on the target dataset.\textbf{Weight only transfer [WT-TGN]} - We train \tgn \cite{tgn} from scratch for 30 epochs on the source dataset and fine-tune the trained weights on the target dataset.\textbf{Memory and weight transfer [\mtgn]} - This is our proposed framework as described in the methodology section, where we use \cite{tgn} as \tgnn.

    
    

\vspace{-0.2cm}
\subsection{Evaluation Metrics} 
\label{app:metrics}
For evaluation, we use metrics including Average Precision (AP), Area under Curve (AUC), Mean Reciprocal Rank (MRR), and Recall@20. \textbf{AP and AUC} are the standard metrics for evaluating \tgnn \cite{tgat,tgn,pint} on the future link prediction task. Positive links are from the validation/test dataset edges, while an equal number of negative links are randomly sampled from non-existent links. We assign label $1$ to positive links and  0 to the negative links. We calculate AP \footnote{scikit-learn: \url{https://scikit-learn.org/stable/modules/generated/sklearn.metrics.average_precision_score.html}} and AUC \cite{AUC} based on link prediction probabilities. \textbf{Mean Reciprocal Rank (\text{MRR})} evaluates rank performance. Given a dataset $D$ of interactions $(u,l,t)$ where  $u$, $l$, and $t$ denote the user, item, and timestamp, respectively. To compute \text{MRR} on an interaction $(u,l,t)$, we generate recommendations for $u$ at timestamp $t$ and compute the rank of ground truth $l$ in the recommended list. MRR over dataset $D$ is $\text{MRR} = \frac{1}{N} \sum_{(u,l,t) \in D} \frac{1}{\text{rank}_{l,u,t}}$. For non-time dependent baselines, $\text{rank}_{l,u,t}$ is the same for all time stamps. \textbf{Recall@20} is the fraction of times the ground truth appears in the top 20 recommendations over the entire validation/test set.  Please note that we report the mean and standard deviation of five runs in the evaluation for each method. Standard deviations smaller than 0.001 are approximated as 0.

\begin{table}[!t]
\vspace{-0.15in}
\caption{Transfer comparison on various pairs from Yelp dataset. We transfer from source cities Philadelphia (PA) and New Orleans (NO) to target cities King of Prussia (KP), Hermitage (HM), Glen Mills (GM) with data scarcity. }
\vspace{-0.15in}
\centering
\resizebox{0.9\columnwidth}{!}{%
\begin{tabular}{llccc}


\toprule
\textbf{Transfer} & \textbf{Method} & \textbf{AP} &\textbf{AUC} &  \textbf{MRR} \\
\midrule
 \multirow{5}{*}{PA $\rightarrow$ KP} & NGCF & $ 0.5389 \pm 0.0 $& $0.5389 \pm 0.0$& $0.0914 \pm 0.0$
 \\
 & LightGCN & $0.5277 \pm 0.0$& $0.5299 \pm 0.0$& $0.0756 \pm 0.0 $ \\

 \cmidrule{2-5}
& NT-TGN &$ 0.5791 \pm 0.0027 $&$ 0.5746 \pm 0.0062 $&$ 0.1219 \pm 0.0012 $\\
& WT-TGN &$ 0.532 \pm 0.0143 $&$ 0.4995 \pm 0.0113 $&$ 0.1085 \pm 0.0139 $\\
& \mtgn &$ \mathbf{0.6567 \pm 0.0015} $&$ \mathbf{0.662 \pm 0.002} $&$ \mathbf{0.1622 \pm 0.0069} $\\

\cmidrule{1-5}

 \multirow{5}{*}{PA $\rightarrow$ HM} & NGCF & $0.5585 \pm 0.0$& $0.5482 \pm 0.0$& $0.05047 \pm 0.0$ \\

 & LightGCN & $0.5172 \pm 0.0$& $0.5031 \pm 0.0$& $0.0518\pm 0.0$ \\

 \cmidrule{2-5}
& NT-TGN &$ 0.6374 \pm 0.0024 $&$ 0.6193 \pm 0.0081 $&$ 0.1005 \pm 0.0061 $\\
& WT-TGN &$ 0.5579 \pm 0.0084 $&$ 0.5431 \pm 0.0081 $&$ 0.0646 \pm 0.0073 $\\
& \mtgn &$ \mathbf{0.6735 \pm 0.0028} $&$ \mathbf{0.6795 \pm 0.0031} $&$ \mathbf{0.1235 \pm 0.0134} $\\

 \cmidrule{1-5}

  \multirow{5}{*}{PA $\rightarrow$ GM} & NGCF & $ 0.5134 \pm 0.0$& $0.5072 \pm 0.0$& $0.0547 \pm 0.0$ \\

 & LightGCN & $0.5206 \pm 0.0$& $ 0.5047\pm 0.0$& $0.05722 \pm 0.0$ \\

 \cmidrule{2-5}
& NT-TGN &$ 0.6052 \pm 0.0073 $&$ 0.602 \pm 0.0066  $&$ 0.0983 \pm 0.0158 $\\
& WT-TGN &$ 0.5721 \pm 0.0029 $&$ 0.5722 \pm 0.0 $&$ 0.0753 \pm 0.0091 $\\
& \mtgn &$ \mathbf{0.6642 \pm 0.0102} $&$ \mathbf{0.6802 \pm 0.0011} $&$ \mathbf{0.1204\pm 0.0023} $\\

  \cmidrule{1-5}

  \multirow{5}{*}{NO $\rightarrow$ KP} & NGCF & $ 0.5389 \pm 0.0 $& $0.5389 \pm 0.0$& $0.0914 \pm 0.0$\\
 & LightGCN & $0.5277 \pm 0.0$& $0.5299 \pm 0.0$& $0.0756 \pm 0.0 $ \\

 \cmidrule{2-5}
& NT-TGN &$ 0.5791 \pm 0.0035 $&$ 0.5746 \pm 0.0061 $&$ 0.1219 \pm 0.0052 $\\
& WT-TGN &$ 0.493 \pm 0.0078 $&$ 0.4638 \pm 0.0077 $&$0.0666 \pm 0.0099 $\\
& \mtgn &$ \mathbf{0.6959 \pm 0.0038} $&$ \mathbf{0.7029 \pm 0.0039} $&$ \mathbf{0.1917 \pm 0.017} $\\

   \cmidrule{1-5}

 \multirow{5}{*}{NO $\rightarrow$ HM} & NGCF & $0.5585 \pm 0.0$& $0.5482 \pm 0.0$& $0.05047 \pm 0.0$ \\

 & LightGCN & $0.5172 \pm 0.0$& $0.5031 \pm 0.0$& $0.0518\pm 0.0$ \\

 \cmidrule{2-5}
& NT-TGN &$ 0.6374 \pm 0.0091 $&$ 0.6193 \pm 0.0154 $&$ 0.1005 \pm 0.0032 $\\
& WT-TGN &$ 0.5542 \pm 0.0058 $&$ 0.5329 \pm 0.0062 $&$ 0.0591 \pm 0.0027 $\\
& \mtgn &$ \mathbf{0.6904 \pm 0.0} $&$ \mathbf{0.6715 \pm 0.002} $& $\mathbf{0.1511 \pm 0.0012} $\\

\cmidrule{1-5}

 \multirow{5}{*}{NO $\rightarrow$ GM} & NGCF & $ 0.5134 \pm 0.0$& $0.5072 \pm 0.0$& $0.0547 \pm 0.0$\\

 & LightGCN & $0.5206 \pm 0.0$& $ 0.5047\pm 0.0$& $0.05722 \pm 0.0$\\

 \cmidrule{2-5}
& NT-TGN &$ 0.6052 \pm 0.0035 $&$ 0.602 \pm 0.0019 $&$ 0.0983 \pm 0.0043 $\\
& WT-TGN &$ 0.5643 \pm 0.0024 $&$ 0.5646 \pm 0.0029 $&$ 0.0779 \pm 0.0049 $\\
& \mtgn &$ \mathbf{0.666 \pm 0.0014} $&$ \mathbf{0.6719 \pm 0.0012} $&$ \mathbf{0.1301 \pm 0.0034} $\\

\bottomrule
\end{tabular}%
}
\label{tab:main_results_yelp}
\vspace{-0.15in}
\end{table}

\begin{table}[h]
\vspace{-0.15in}
\caption{Transfer comparison on various pairs from Wiki-Edits dataset. We choose English (en), French (fr), German (de) as sources and Norwegian (nn), Japanese (ja), Vietnamese (vi), Slovak (sk), Serbian (sr), Arabic (ar) as target languages with data scarcity.}
\vspace{-0.15in}
\centering
\resizebox{0.9\columnwidth}{!}{%
\begin{tabular}{llccc}


\toprule
\textbf{Transfer} & \textbf{Method} & \textbf{AP} &\textbf{AUC} &  \textbf{MRR}\\
\midrule
 \multirow{5}{*}{en $\rightarrow$ nn} & NGCF & $ \mathbf{0.6823 \pm 0.0}  $& $0.6243 \pm 0.0 $& $0.2716 \pm 0.0 $
 \\
 & LightGCN & $0.6360 \pm 0.0 $& $0.5598 \pm 0.0 $& $0.2349 \pm 0.0  $\\

 \cmidrule{2-5}
& NT-TGN &$ 0.5872 \pm 0.0198  $&$ 0.5996 \pm 0.0234  $&$ 0.0919 \pm 0.0045  $\\
& WT-TGN &$ 0.4918 \pm 0.0106  $&$ 0.4153 \pm 0.0130  $&$ 0.1415 \pm 0.0041  $\\
& \mtgn &$ 0.6733 \pm 0.0148  $&$ \mathbf{0.6991 \pm 0.0200 } $&$ 0.2819 \pm 0.0088  $\\

\cmidrule{1-5}

 \multirow{5}{*}{en $\rightarrow$ ja} & NGCF & $0.7043 \pm 0.0 $& $0.6547 \pm 0.0 $& $0.1264 \pm 0.0 $ \\

 & LightGCN & $0.6441 \pm 0.0 $& $0.5821 \pm 0.0 $& $0.1207 \pm 0.0 $ \\

 \cmidrule{2-5}
& NT-TGN &$ 0.5402 \pm 0.0135  $&$ 0.4930 \pm 0.0165  $&$ 0.0225 \pm 0.0004  $\\
& WT-TGN &$ 0.7589 \pm 0.0295  $&$ 0.7148 \pm 0.0180  $&$ 0.1165 \pm 0.0050  $\\
& \mtgn &$ \mathbf{0.8673 \pm 0.0415 } $&$ \mathbf{0.8657 \pm 0.0303 } $&$ \mathbf{0.1961 \pm 0.0084 } $\\






  


   \cmidrule{1-5}

 \multirow{5}{*}{fr $\rightarrow$ sk} & NGCF & $0.5244 \pm 0.0 $& $0.4600 \pm 0.0 $& $0.0556 \pm 0.0 $\\
 
 & LightGCN & $0.5465 \pm 0.0 $& $0.4604 \pm 0.0 $& $0.0572 \pm 0.0 $\\

 \cmidrule{2-5}
& NT-TGN &$ 0.4943 \pm 0.0244  $&$ 0.4669 \pm 0.0074  $&$ 0.0361 \pm 0.0016  $\\
& WT-TGN &$ 0.5689 \pm 0.0280  $&$ 0.4536 \pm 0.0054  $&$ 0.0562 \pm 0.0018  $\\
& \mtgn &$ \mathbf{0.7760 \pm 0.0085 } $&$ \mathbf{0.7408 \pm 0.0237} $& $\mathbf{0.1221 \pm 0.0034 } $\\

\cmidrule{1-5}

 \multirow{5}{*}{fr $\rightarrow$ sr} & NGCF & $ 0.5454 \pm 0.0 $& $0.5243 \pm 0.0 $& $0.0355 \pm 0.0 $\\
 & LightGCN & $0.5238 \pm 0.0 $& $0.5096 \pm 0.0 $& $0.0323 \pm 0.0 $ \\

 \cmidrule{2-5}
& NT-TGN &$ 0.4840 \pm 0.0158  $&$ 0.4900 \pm 0.0138  $&$ 0.0133 \pm 0.0004  $\\
& WT-TGN &$ 0.5863 \pm 0.0086  $&$ 0.5245 \pm 0.0240  $&$ 0.0596 \pm 0.0029  $\\
& \mtgn &$ \mathbf{0.6926 \pm 0.0204 } $&$ \mathbf{0.6617 \pm 0.0075 } $&$ \mathbf{0.0749 \pm 0.0016 } $\\

\cmidrule{1-5}

 \multirow{5}{*}{de $\rightarrow$ ar} & NGCF & $ 0.6357 \pm 0.0 $& $0.5905 \pm 0.0 $& $0.0725 \pm 0.0 $ \\
 & LightGCN & $0.6299 \pm 0.0 $& $0.5882 \pm 0.0 $& $0.0976 \pm 0.0 $ \\

 \cmidrule{2-5}
& NT-TGN &$ 0.5339 \pm 0.0104  $&$ 0.5018 \pm 0.0116  $&$ 0.0198 \pm 0.0010  $\\
& WT-TGN &$ 0.7050 \pm 0.0275  $&$ 0.6722 \pm 0.0083  $&$ 0.0860 \pm 0.0025  $\\
& \mtgn &$ \mathbf{0.7602 \pm 0.0271 } $&$ \mathbf{0.7602 \pm 0.0268 } $&$ \mathbf{0.1481 \pm 0.0027 } $\\




\cmidrule{1-5}

 \multirow{5}{*}{de $\rightarrow$ sk} & NGCF & $0.5244 \pm 0.0 $& $0.4600 \pm 0.0 $& $0.0556 \pm 0.0 $\\
 
 & LightGCN & $0.5465 \pm 0.0 $& $0.4604 \pm 0.0 $& $0.0572 \pm 0.0 $ \\

 \cmidrule{2-5}
& NT-TGN &$ 0.4943 \pm 0.0109  $&$ 0.4669 \pm 0.0162  $&$ 0.0361 \pm 0.0009  $\\
& WT-TGN &$ 0.5474 \pm 0.0140  $&$ 0.4908 \pm 0.0078  $&$ 0.0612 \pm 0.0017  $\\
& \mtgn &$ \mathbf{0.7067 \pm 0.0189 } $&$ \mathbf{0.6978 \pm 0.0267 } $&$ \mathbf{0.1226 \pm 0.0034 } $\\

\bottomrule
\end{tabular}%

}
\label{tab:main_results_wiki}
\vspace{-0.15in}
\end{table}

\vspace{-0.1in}
\subsection{Empirical Evaluation}

\begin{figure}[h]
    \centering
        \includegraphics[scale=0.41]{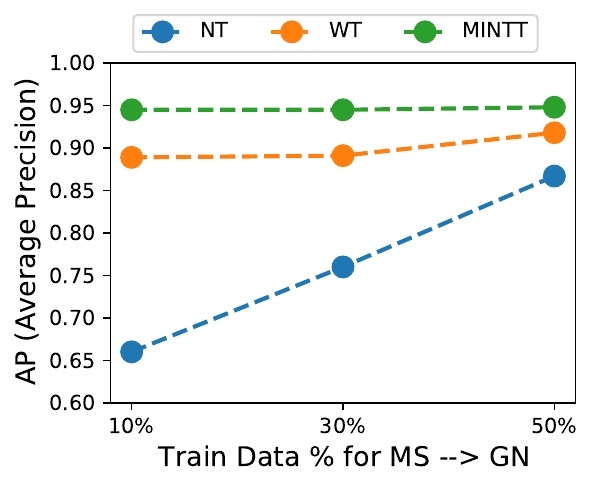}
        \includegraphics[scale=0.41]{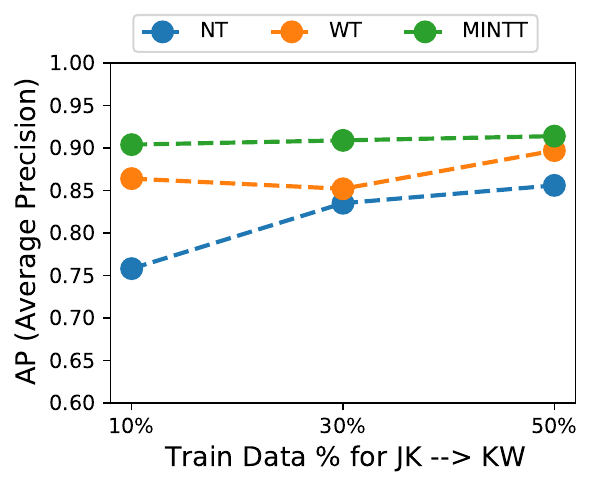}
    \vspace{-0.1in}
    \caption{Performance of proposed framework \namemodelnew in different extents of train data scarcity on Foursquare}
    \label{fig:mint_data_scarcity}
    \vspace{-0.1in}
\end{figure}

We now discuss the three key questions posed at the start of the section. We extensively evaluate the performance of our transfer framework and compare it to various baselines and our method variants across different source regions in foursquare, i.e., Moscow (MS), Jakarta (JK), Tokyo (TK), and Kuala Lumpur (KL). We show results on transfers to several target regions, i.e., Gent (GN), Brussels (BR), Kuwait (KW), Rio de Janeiro (RJ), Campo Grande (CG), and San Antonio (SA). Table \ref{tab:main_results5} shows the performance comparison of \namemodelnew with baselines and \namemodelnew-variants with scarcity simulating training conditions on target graphs as described in section 5.2. We indeed observe that the proposed method \namemodelnew works considerably well compared to baselines and its variants by a margin of 30\%-60\% in AP. Improvements in other metrics, \text{MRR} and Recall@20, are much better. 

We observe that the weight-only transfer variant WT-TGN considerably outperforms the no-memory and no-weight transfer variant NT-TGN. The proposed framework \namemodelnew consistently outperforms the weight-only transfer variant WT-TGN across all metrics, highlighting the explicit advantage of the memory transfer framework.  Moreover, table \ref{tab:main_results_tgat_gm} underscores the benefit of memory transfer in a data-scarce setting in comparison to non-memory based \tgnn. While non-memory \tgnn, \tgat, and \gm do better than no-weight and no-memory transfer \tgn on the target graph, their weight-transfer variants WT-\tgat and WT-\gm performs worse than \namemodelnew. This evaluation undoubtedly strengthens the efficacy of memory transfer in a data-scarce target graph. We For similar evaluation on Wiki and Yelp, we refer the reader to tables \ref{tab:main_results_yelp} and \ref{tab:main_results_wiki}. This concludes the evaluation on questions \textbf{RQ1} and \textbf{RQ2}. 

We notice that among baselines, \ngcf and \lightgcn, the latter outperforms the former. This is expected as \lightgcn is a simpler model less prone to over-fitting.  It does much better in \textsc{MRR} and recall@20.  In some instances, \lightgcn does better than NT-TGN owing to the poor performance of \tgnn in scarce data settings.

Table \ref{tab:main_results_6} summarizes the average performance for all source datasets across all 79 target datasets. We observe a similar trend, i.e., \mtgn outperforms other variants. Figures \ref{fig:mint_evalution_moscow_tokyo} compare transfer performance from source graphs (Tokyo, Moscow, Kuala Lumpur, and Jakarta) to multiple other target regions. Here, we also observe similar trends.


\begin{figure}[t]
    \centering
    \begin{subfigure}{\linewidth}
        \centering
        \includegraphics[scale=0.18]{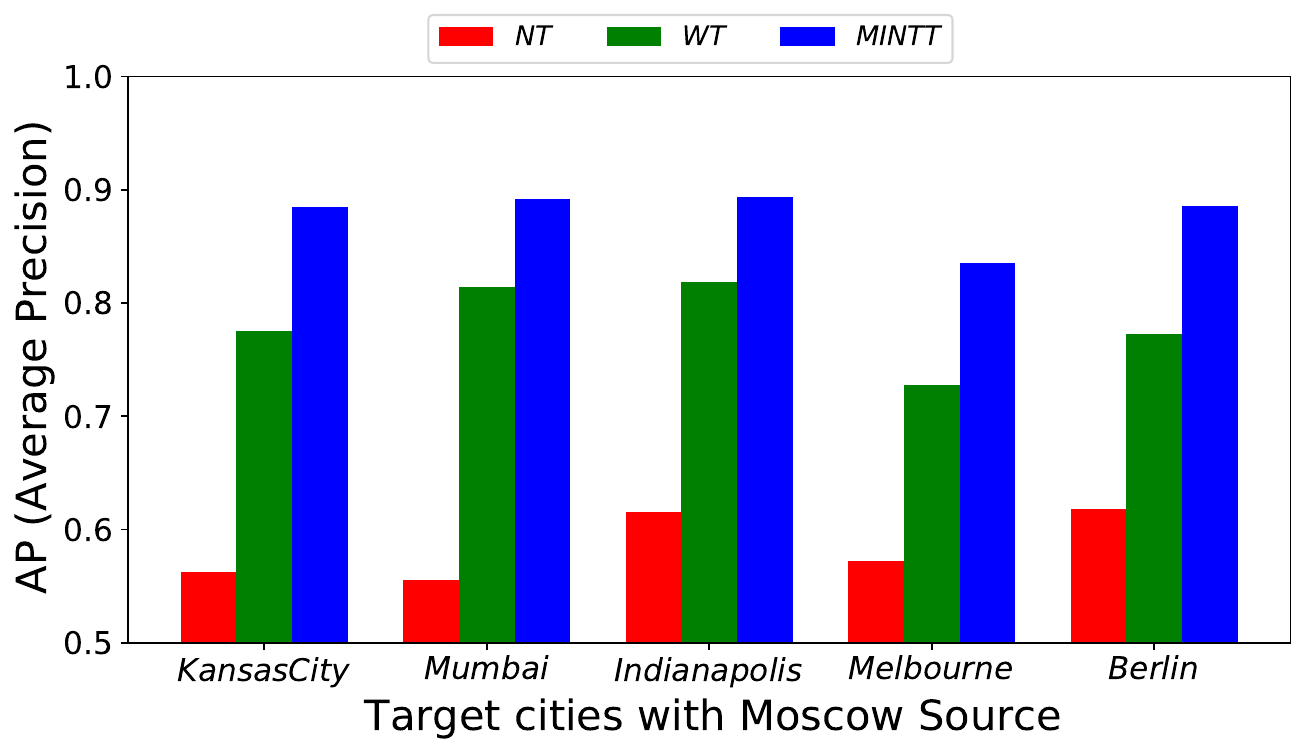}
        \includegraphics[scale=0.18]{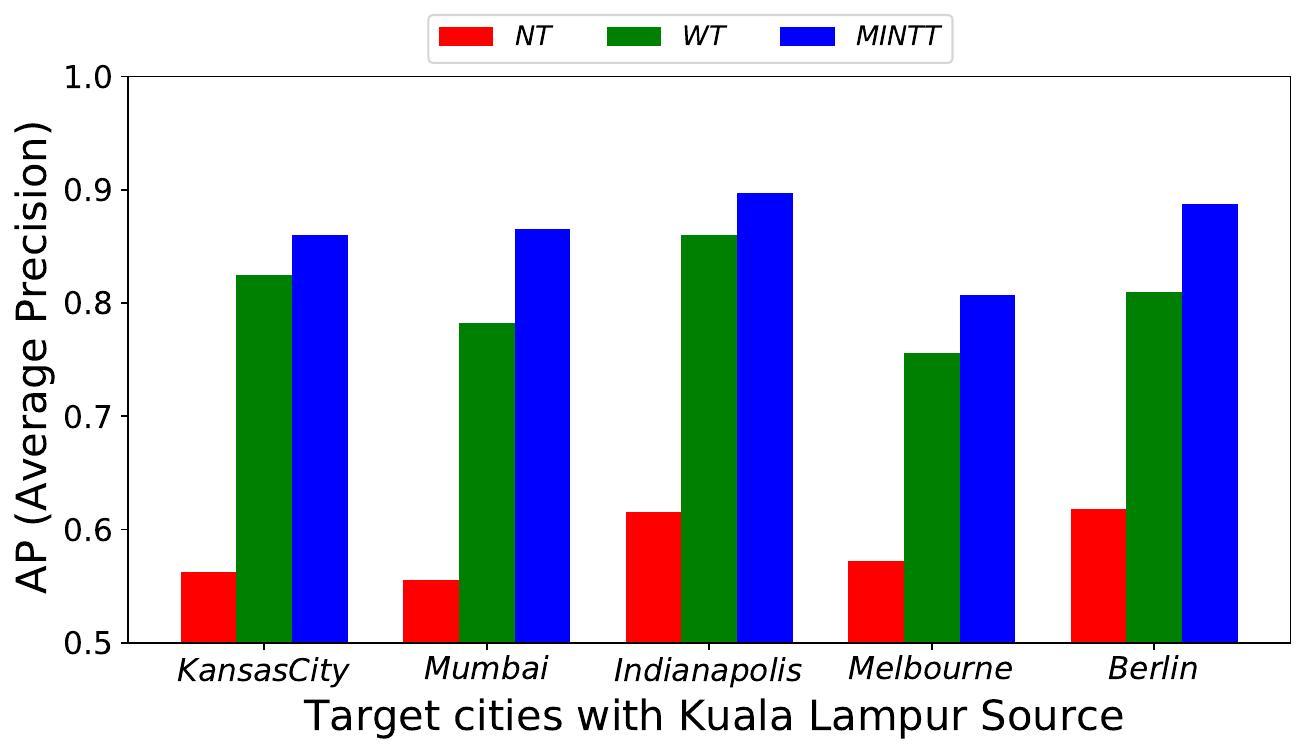}
    \end{subfigure}
    \begin{subfigure}{\linewidth}
        \centering
        \includegraphics[scale=0.18]{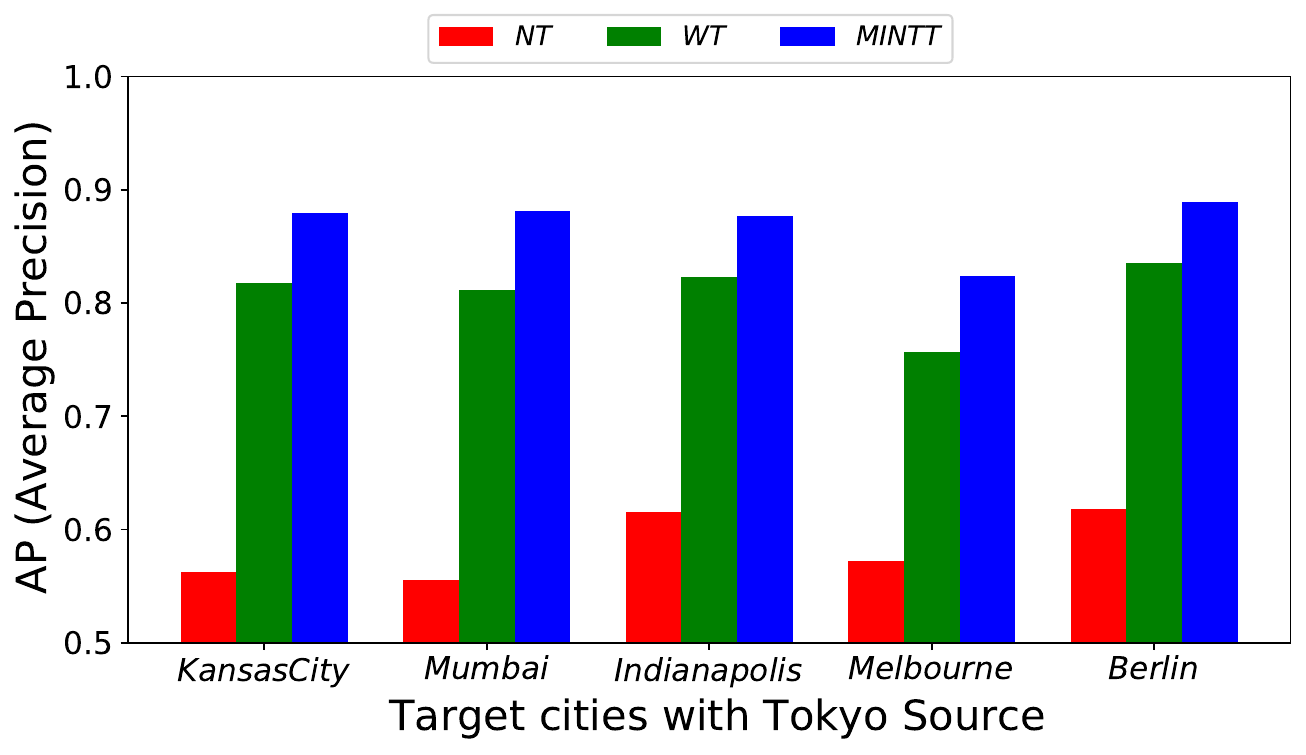}
        \includegraphics[scale=0.18]{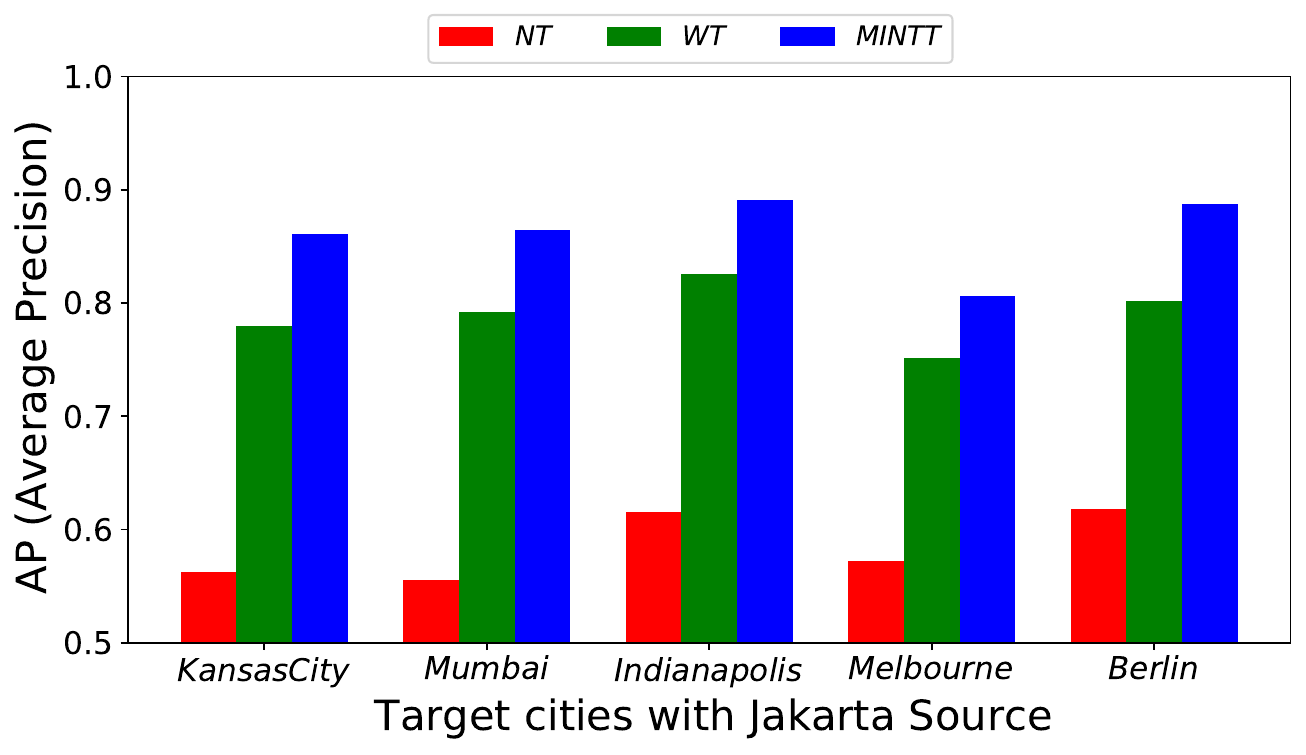}        
        \vspace{-0.1in}
    \end{subfigure}
    \vspace{-0.2in}
    \caption{Visualization of our transfer framework's performance across different source-target region pairs.}
    \label{fig:mint_evalution_moscow_tokyo}
    \vspace{-0.2in}
\end{figure}

\begin{table}[h]
\vspace{-0.1in}
\caption{AP and MRR of \namemodelnew and its variants on different extents of training data scarcity on Foursquare}
\vspace{-0.15in}
\centering
\resizebox{0.9\columnwidth}{!}{%
\begin{tabular}{c|c|cc|cc|cc}



\toprule
\textbf{Transfer} & \textbf{Method} & \multicolumn{2}{c}{\textbf{Target 50\%}}  &\multicolumn{2}{c}{\textbf{Target 30\%}}  &  \multicolumn{2}{c}{\textbf{Target 10\%}} \\
\midrule

\multirow{4}{*}{MS $\rightarrow$ GN} & &   \textbf{AP} & \textbf{MRR} & \textbf{AP} & \textbf{MRR} & \textbf{AP} & \textbf{MRR} \\
\cmidrule{3-8}
& NT-TGN & 0.8672 & 0.1088 & 0.7606 & 0.044 & 0.6666 & 0.0254 \\
& WT-TGN & 0.9186 & 0.2856 & 0.8915 & 0.1964 & 0.8892 & 0.1665 \\
& MWT-TGN & 0.9486 & 0.3818 & 0.9459 & 0.3443 & 0.9459 & 0.3757 \\
\cmidrule{1-8}

\multirow{3}{*}{JK $\rightarrow$ KW} 
& NT-TGN & 0.8569 & 0.1441 & 0.8359 & 0.0938 & 0.7581 & 0.0444 \\
& WT-TGN & 0.8979 & 0.2607 & 0.8527 & 0.1286 & 0.8646 & 0.2096 \\
& MWT-TGN & 0.9145 & 0.3282 & 0.9097 & 0.3038 & 0.9044 & 0.2994 \\
\cmidrule{1-8}

\multirow{3}{*}{TY $\rightarrow$ SA}
& NT-TGN & 0.8291 & 0.0967 & 0.8280 & 0.0758 & 0.6737 & 0.0180 \\
& WT-TGN & 0.8499 & 0.1911 & 0.8578 & 0.1227 & 0.8127 & 0.1946 \\
& MWT-TGN & 0.8876 & 0.2529 & 0.8854 & 0.2520 & 0.8723 & 0.2286 \\

\bottomrule
\end{tabular}%
}
\label{tab:main_results_data_scarcity_various}
\vspace{-0.2in}
\end{table}

Finally, to answer \textbf{RQ3}, we evaluate our transfer framework at different extents of data scarcity. We select a few source and target pairs and evaluate the transfer at different training data percentages (50\%,30\%,10\%)  of the target graph while utilizing 20\% data for validation and 30\% for testing. In figure \ref{fig:mint_data_scarcity} and table \ref{tab:main_results_data_scarcity_various}, we observe that in a data-abundant setting (50\% training data), the performance of no-memory and no-weight transfer \tgn (NT-TGN) is comparable with weight-only transfer \tgn (WT-TGN) and \mtgn. However, as data scarcity increases, the performance of NT-TGN and WT-TGN deteriorates rapidly. In contrast, \mtgn consistently upholds its performance consistently, which establishes its efficacy for low-resource target dataset adaptation. This also implies that \textit{\mtgn is not required in a non-scarce setting where the weight-only transfer itself is sufficient}. This concludes our evaluation of the proposed framework \namemodelnew. 
\vspace{-0.2cm}
\section{Discussion on non-memory based \tgnn}
\label{sec:additional_baseline}
We have compared our proposed transfer approach with the transfer of non-memory-based \tgnn like \tgat \cite{tgat} and state-of-the-art \gm \cite{gm} in table \ref{tab:main_results_tgat_gm} highlighting the importance of memory in model transfer. We also scrutinized \dygformer \cite{dygformer}, a non-memory \tgnn outside the scope of this work, as it predicts the link formation probability using link-based features instead of learning node representations like \tgat,\tgn, and \gm. \dygformer's architecture utilizes common neighbor encodings to predict link formation probability, leading to more expressivity \cite{chamberlain2023graph}, which leads to better performance than node representation learning-based \tgnn.  However, unlike node-representation learning-based models that provide dynamic embeddings applicable to multiple downstream tasks like future link prediction, node classification, clustering, and anomaly detection,  \dygformer is computationally expensive for large-scale recommendation tasks as it computes edge-level features, not node embeddings. For a dynamic graph with $N$ nodes, \dygformer requires $O(N^2)$ complexity for future link predictions, whereas \tgn,\tgat and \gm using approximate neighbor indexing \cite{malkov2018efficient} of embeddings can achieve a scalable $O(N{\log}N)$ complexity for computing cosine similarity between embeddings.


\vspace{-0.1in}
\section{Detailed Related Work}
\label{app:related_works}
In this section, we provide more details on the current literature related to transfer learning on graphs.
\cite{kooverjee2022investigating} shows the efficacy of a simple weight transfer by fine-tuning a source graph trained \gnn model on the target graph. \cite{gfewshot} jointly trains \gnn on both the source and target graph with a soft coupling penalty to align the embedding spaces between the source and target graph. \cite{10.5555/3367243.3367352} solves the joint task of link prediction and network alignment using seed nodes mapping to jointly train on both the source and target graph containing the same users, but without known mapping, boosting performance. \cite{srcreweights} proposes a meta-learning-based solution for transfer learning on traffic graphs by dividing source graphs into multiple regions spatially, then learning these region's weight and fine-tuning on target graph. \cite{yao2020learning} extends this to transfer from multiple traffic graphs to the target graph instead of one-to-one transfer. \cite{grl} solves few-shot classification by utilizing the auxiliary graphs containing enough labels per class by learning joint metric space for labels. \cite{cold_start_gnn,pre_train_gnn_cold_start,mamo} are cold-start meta-learning-based recommendation models that quickly adapt to new users and items having few connections in the same graph by simulating cold-start during model training. Unlike our problem \ref{prob:1}, which focuses on transferring knowledge from a trained source \tgnn to the target graph, these methods either focus on joint training of source and target graph models or fine-tuning for new nodes in the source graph itself.

\vspace{-0.1in}
\section{Limitation and Future Work}
\label{sec:future_work}
 
\textbf{Graph Similarity and performance correlation:}
This work does not explore the correlation between source-target graph similarity and performance improvement. It requires elaborate research and remains for future work as similarity metrics for temporal graphs are complex. Thus, we have provided the aggregated results of \tgn transfer from source to all target graphs in table \ref{tab:main_results_6}.

\noindent \textbf{Dynamic node attributes and scalability:} Our work assumes static node features as all open-source temporal graphs lack dynamic attributes. We see this as potential future work.

\noindent \textbf{Handling of continuous node attributes:}
The proposed graph transformation assumes categorical node attributes on both source and target graph datasets. It can be extended to continuous attributes by discretizing them into bins and utilizing these bins as new categories as input to the \namemodelnew. 

\vspace{-0.1in}
\section{Conclusion}
Temporal Graph Neural Networks (\tgnn) have exhibited state-of-the-art performance in future link prediction tasks for temporal interaction graphs. However, they perform poorly in low data settings, and the node-specific memory modules of \tgnn make them non-trivial for transferring inductive knowledge from semantically similar other graphs. Towards this, we formulate a novel transfer framework \mtgn that learns a memory mapping between the source and target graphs to enable the transfer of \tgnn memory to enhance performance on target graphs. We evaluate the performance by transferring \tgnn across several sources and data-scarce target graph pairs extracted from a real-world dataset spanning 400 regions. \mtgn significantly outperforms the \gnn based recommendation baselines suitable for low-data settings, transfer of non-memory based \tgnn, no-memory and no-weight transfer of \tgn,  weight-only transfer of \tgn, establishing the importance of \tgnn memory transfer in low-resource data setting. In terms of future work, we wish to explore meta-learning-based solutions for fine-tuning the trained \tgn model on target graphs by simulating the data scarcity on the source graph. We note this is a different set-up with respect to this work, as the source \tgnn model is already trained, and we leverage it to fine-tune the model on the target graph. \looseness=-1

\bibliographystyle{ACM-Reference-Format}
\bibliography{references}









\end{document}